\PassOptionsToPackage{table,xcdraw,dvipsnames}{xcolor}

\documentclass[11pt,letterpaper]{plum}

\usepackage[all]{hypcap}
\PassOptionsToPackage{round}{natbib}
\usepackage{natbib}
\usepackage{nicefrac}
\usepackage{arydshln}
\usepackage{subcaption}
\usepackage{pgfplots}
\pgfplotsset{compat=1.18}
\usepgfplotslibrary{groupplots}
\tcbuselibrary{most,skins,theorems,breakable,xparse}
\usepackage{listings}
\usepackage{xspace}
\usepackage{soul}
\usepackage{cleveref}
\usepackage{etoc}
\usepackage[section]{placeins}

\usepackage{algorithm}
\usepackage{algpseudocode}

\algrenewcommand\alglinenumber[1]{\scriptsize #1}

\usepackage{multirow}
\usepackage{makecell}
\usepackage{wrapfig}
\usepackage{float}
\usepackage{stackengine}
\usepackage{mathtools}
\usepackage{array}

\definecolor{darkblue}{rgb}{0,0,0.5}
\definecolor{lightblue}{RGB}{220,235,250}
\definecolor{tabhead}{HTML}{DCE6F1}
\definecolor{tabband}{HTML}{F2F2F2}
\definecolor{cellgood}{HTML}{D8ECD2}
\definecolor{cellbad}{HTML}{F4D2D2}
\definecolor{cellbase}{HTML}{E8E8E8}
\definecolor{textgood}{HTML}{1B7A2D}
\definecolor{textbad}{HTML}{B22222}
\definecolor{textmute}{HTML}{6B6B6B}

\definecolor{bg}{gray}{0.95}
\definecolor{lightgray}{gray}{0.95}
\definecolor{plum}{RGB}{200,140,240}
\definecolor{lavender}{RGB}{180,160,255}
\definecolor{deepteal}{HTML}{004953}
\definecolor{headergray}{HTML}{F5F5F5}

\definecolor{pearbg}{HTML}{fff2ea}
\definecolor{pearfg}{HTML}{002a3a}
\definecolor{algorange}{HTML}{e45a07}

\definecolor{purplelight}{RGB}{245,240,255}
\definecolor{purplemid}{RGB}{140,95,200}

\definecolor{lightorange}{HTML}{faa755}
\definecolor{annothl}{RGB}{255,232,150}
\definecolor{annotaccent}{RGB}{170,40,30}
\definecolor{diffaccent}{RGB}{30,90,170}
\sethlcolor{annothl}

\hypersetup{colorlinks=true,citecolor=darkblue,linkcolor=darkblue,urlcolor=darkblue,
  pdftitle={Good SFT Optimizes for SFT, Better SFT Prepares for Reinforcement Learning},
  pdfauthor={Dylan Zhang, Yufeng (Felix) Xu, Haojin Wang, Qingzhi Chen, Hao Peng}}

\newcolumntype{P}{>{\columncolor{pearbg}}c}

\newcommand{\PassFont}{\sffamily\bfseries}
\newcommand{\Pass}[1]{{\PassFont #1}}

\theoremstyle{plain}

\theoremstyle{definition}

\theoremstyle{remark}

\newcommand{\name}{PEAR\xspace}
\newcommand{\tgt}{$\pi_\theta$\xspace}
\newcommand{\beh}{$\pi_\beta$\xspace}
\newcommand{\namebasic}{PEAR{\tiny{B=1}}}
\newcommand{\fullname}{\textbf{P}olicy \textbf{E}valuation--inspired \textbf{A}lgorithm for Offline Learning Loss \textbf{R}eweighting}

\usepackage[textsize=tiny]{todonotes}

\newtcolorbox{takeawaybox}[1][\textsc{Takeaways}]{%
  enhanced,
  colback=white,
  colframe=deepteal,
  boxrule=0.6pt,
  arc=1.5mm,
  left=1.2mm,right=1.2mm,top=1.0mm,bottom=1.0mm,
  fonttitle=\bfseries,
  title={#1}
}
\newtcolorbox{rqbox}{%
  enhanced,
  colback=deepteal!4,
  colframe=deepteal,
  boxrule=0.6pt,
  arc=1.5mm,
  left=1.2mm,right=1.2mm,top=1.0mm,bottom=1.0mm
}
\newtcolorbox{namesummarybox}[1][\textsc{Approach Summary}]{%
  enhanced,
  colback=white,
  colframe=deepteal,
  boxrule=0.6pt,
  arc=1.5mm,
  left=1.2mm,right=1.2mm,top=1.0mm,bottom=1.0mm,
  fonttitle=\bfseries,
  title={#1}
}

\newcounter{rq}
\newcommand{\RQ}[1][]{%
  \refstepcounter{rq}%
  \textbf{RQ}\,\arabic{rq}\xspace%
  \if\relax\detokenize{#1}\relax\else\label{#1}\fi%
}

\usepackage{amsmath,amsfonts,bm}

\def\1{\bm{1}}

\DeclareMathAlphabet{\mathsfit}{\encodingdefault}{\sfdefault}{m}{sl}
\SetMathAlphabet{\mathsfit}{bold}{\encodingdefault}{\sfdefault}{bx}{n}

\newcommand{\algc}[1]{\textcolor{algorange}{{\hfill{\footnotesize$\triangleright$~#1}}}}

\title{Good SFT Optimizes for SFT, Better SFT Prepares for Reinforcement Learning}

\author[1]{Dylan Zhang}
\affil[1]{University of Illinois Urbana-Champaign}
\author[1,2]{Yufeng (Felix) Xu}
\affil[2]{New York University (Shanghai). Work done during internship at UIUC.}
\author[1]{Haojin Wang}
\author[1]{Qingzhi Chen}
\author[1]{Hao Peng}
\correspondingauthor{Dylan Zhang, \href{mailto:shizhuo2@illinois.edu}{shizhuo2@illinois.edu}; Hao Peng, \href{mailto:haopeng@illinois.edu}{haopeng@illinois.edu}}

\begin{abstract}
Post-training of reasoning LLMs is a holistic process that typically consists of an offline SFT stage followed by an online reinforcement learning (RL) stage.
However, SFT is often optimized in isolation to maximize SFT performance alone.
We show that,
after identical RL training, models initialized from stronger SFT checkpoints can significantly underperform those initialized from weaker ones.
We attribute this to a mismatch typical in current SFT--RL pipelines: the distribution that generates the offline SFT data can differ substantially from the policy optimized during online RL, which learns from its own rollouts.
We propose \name (\fullname), an SFT-stage method that corrects this mismatch and better prepares the model for RL.
\name uses importance sampling to reweight the SFT loss, with three variants operating at the token, block, and sequence levels.
It can be used to augment standard SFT objectives
and incurs little additional training overhead once probabilities for the offline data are collected.
We conduct controlled experiments on verifiable reasoning games and mathematical reasoning tasks on Qwen2.5/3 and DeepSeek-distilled models.
\name consistently improves post-RL performance over canonical SFT, with Pass@8 gains up to $30$ percentage points on AIME-2025.
Our results suggest that \name is an effective step toward more holistic LLM post-training by designing and evaluating SFT with downstream RL in mind rather than in isolation.
\end{abstract}

\begin{document}

\maketitle
\makeatletter
\@thanks
\let\@thanks\@empty
\makeatother

\section{Introduction}

Post-training of reasoning language models typically follows a two-stage paradigm: an \emph{offline} supervised fine-tuning (SFT) phase produces an initial checkpoint, which is then used to initialize an 
\emph{online} reinforcement learning (RL) phase that further enhances the model~\citep{shao2024deepseekmath,Guo_2025,yang2024qwen25}. Both areas have become active research fronts. In particular, a growing body of work
has proposed offline learning objectives to improve SFT,
often by reweighting or regularizing next-token likelihood~\citep{qin2025iwsft,zhu2025proximalsft,wu2025dft,lin2025sftdoesnthurtgeneral,li2025beyond}.

\begin{figure*}
\centering
\includegraphics[width=\linewidth]{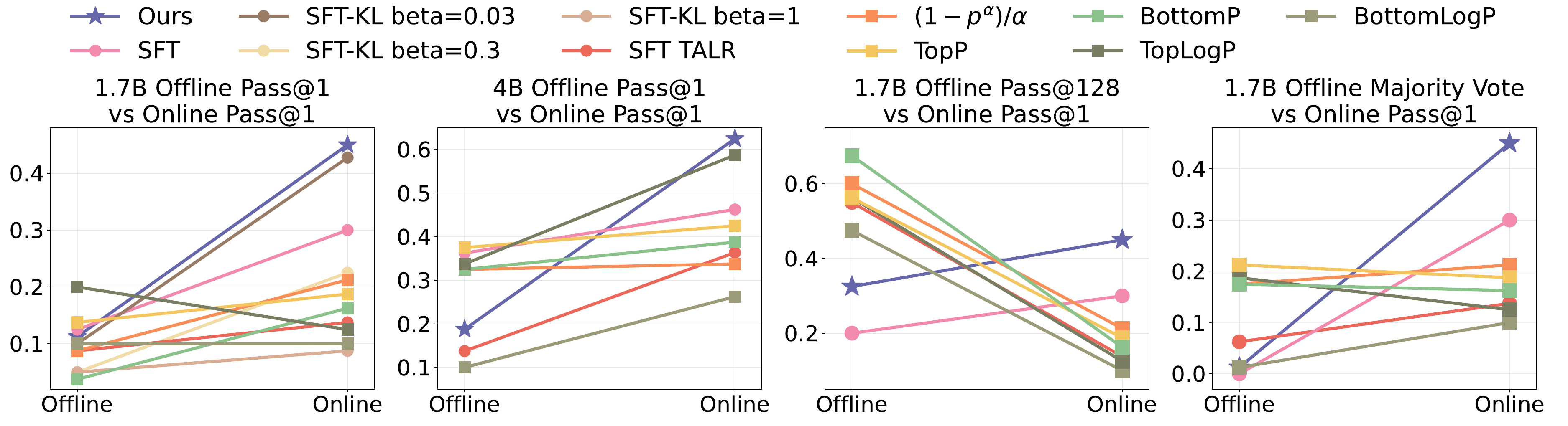}
    \caption{Offline v.s. Online pass@1 on SynLogic Games on a total of 19 Models. It exhibits significant ranking changes indicating offline performance will not entail online performance. In addition, our proposed approach remains the most effective initialization for online RL.}
    \label{fig:offline_online}
\end{figure*}

From a practical perspective, the performance of interest is usually the model's final accuracy after completing both SFT and downstream RL.
However, it is common that these existing techniques optimize for SFT-stage performance in isolation,
often with the implicit assumption that gains in offline performance will translate to improved performance after RL. \citet{kang2025quagmiressftrlposttraininghigh} show that repetition and data homogeneity boost SFT but may reduce RL headroom. 
This motivates us to investigate if offline gains of an objective could also be a misleading proxy for its effectiveness as an RL initialization. We empirically show the gains of a stronger offline checkpoint over a weaker one can shrink, disappear, or even reverse after both undergo identical RL training. Therefore, optimizing for offline performance alone may be counterproductive when the goal is strong final performance after RL (Fig.~\ref{fig:offline_online} in \S\ref{sec:offline_not_entail_online}).


We contend that the goal of an offline stage is not merely strong offline accuracy, but an initialization that facilitates improvement under the online RL. 
This requires addressing a distribution mismatch between offline and online stages:
Typically, during SFT, the model learns from data sampled from a different distribution, often dubbed the \textbf{behavior policy}~\citep{sutoonrl,pdis2000,uehara2022opesurvey}.
In contrast, during online RL, the target of learning (thereby the \textbf{target policy}), learns from roll-outs generated by itself. There is a clear distribution mismatch that needs to be corrected between them ~\citep{zhao2022adaptivebehaviorcloningregularization,lee2021offlinetoonlinereinforcementlearningbalanced,zu2025behavioradaptiveqlearningunifyingframework} in order for an effective offline-to-online transition. 

It is therefore crucial to quantify and correct this distribution mismatch. 
Inspired by off-policy evaluation (OPE)~\citep{pdis2000,thomas2016dataefficientoffpolicypolicyevaluation,jiang2016doublyrobustoffpolicyvalue,levine2020offlinereinforcementlearningtutorial},
we address this by reweighting offline data using importance weights, i.e., the likelihood ratio between the target policy and the behavior policy (\S\ref{sec:method}).
Intuitively, this reweighting scales each token's loss according to how likely the target policy would generate the same continuation relative to the behavior policy, so that offline training better reflects the trajectories that online RL will actually revisit. We present a sequence-level ~\citep{cis_off_policy} and a token-wise reweighting based on suffix-ratios~\citep{pdis2000}. We also present variants that improve stability by block-wise weighing and leveraging negative data. 


We evaluate \name and its variants on reasoning games and math benchmarks across 6 different models of various sizes: Qwen3-Base-0.6B; 1.7B; 4B and 8B~\citep{yang2025qwen3technicalreport}, Qwen2.5-1.5B-Math~\citep{yang2024qwen25} and DeepSeek-Distill-Qwen-1.5B~\citep{Guo_2025}. 
Using an SFT–RL pipeline that varies only the SFT-stage objectives, \name and its variants consistently improve \emph{post-RL} performance over strong baselines.
Comparing checkpoints finetuned using \name versus canonical SFT on the same data,
the former outperforms the latter by up to $+25$ pp absolute Pass@1 on SynLogic logic games (Qwen3-1.7B-Base, $13.1\%\to 38.3\%$; Table~\ref{tab:synlogic_enigmata})
and up to $+30$ pp Pass@8 on AIME-25~\citep{matharena} (DS-Qwen-1.5B, $5\%\to 35\%$; Table~\ref{tab:math_results}).
Moreover, our analysis shows that \name-initialized models undergo less parameter drift during RL~(\S\ref{sec:analysis}). The takeaway is simple: \emph{a good offline stage should not optimize for offline accuracy, but prepare the policy for the RL that follows}. \name is one concrete instantiation of this principle.


\section{Offline Performance May Not Entail Online}
\label{sec:offline_not_entail_online}


There are various techniques to improve supervised fine-tuning (SFT) for reasoning, typically targeting stronger \emph{offline} performance or reduced forgetting.

Recent reasoning LM post-training pipeline typically applies an online RL stage to further improve performance after SFT~\citep{shao2024deepseekmath,Guo_2025,yang2024qwen25}.
In this setting, the offline stage provides the initialization for RL, and prior study~\citep{kang2025quagmiressftrlposttraininghigh} has identified that dataset construction and hyper-parameter affects SFT and RL performances differently in this pipeline. This naturally leads us to a question: 

\begin{rqbox}
Will the advantage of an offline learning objective carry over to post-RL performance?
\end{rqbox}

We experiment with a wide spectrum of objectives covering the span the standard SFT ``loss-strength'' spectrum: drift control (KL), smooth probability-shaped reweighting and hard masking toward high-/low-confidence tokens. 

\citet{li2025beyond} presents a generalized view of SFT loss by studying a series of probability-based objectives (Table~\ref{tab:objective_name_fn}), where one could control how strongly training emphasizes low- vs.\ high-probability tokens by altering the transformation of probability, and learning selectively from easy / difficult tokens. TALR similarly modifies SFT via adaptive token-wise reweighting (Table~\ref{tab:objective_name_fn}). We follow their recommended hyperparameters; see Appendix~\ref{app:bll-hparams} and ~\ref{app:talr_hparams}. 

In addition, we consider standard negative log-likelihood (NLL) loss; and KL-regularized NLL \(\mathcal{L}(\theta)
=
\mathbb{E}_{(x,y)\sim\mathcal{D}}\!\big[-\log \pi_\theta(y\mid x)\big]
\;+\;
\beta\,\mathbb{E}_{x\sim\mathcal{D}}\!\left[\mathrm{KL}\!\left(\pi_\theta(\cdot\mid x)\,\|\,\pi_{\rm ref}(\cdot\mid x)\right)\right].\) We alter $\beta \in\{0.03,0.1,0.3,1\}$ for KL-regularized variant.

We perform a controlled, contamination-free experiment by applying each of these offline objectives followed by online RL on synthetic logic puzzles from~\citep{liu2025synlogic}. 

\subsection{Offline $\ne$ Online}
\begin{takeawaybox}
\textbf{Offline $\neq$ online:} better algorithm in offline scores need not yield better post-RL performance.
\end{takeawaybox}
Figure~\ref{fig:offline_online} visualizes offline versus online performance.  
While several objectives indeed outperform SFT offline on Pass@1, these gains do not
reliably translate to stronger post-RL models: some checkpoints are simply harder for subsequent RL to
improve and ultimately lose their offline advantage. One may be tempted to pick \textbf{TopLogP} for Qwen3-1.7B-Base because of the best offline scores, yet this choice leads to worst-among-all post-RL performance, even under-performing SFT initialized model. 

We inspect other descriptors for sampling that potentially relates to RL: offline pass@K with large K~\citep{yue2025doesreinforcementlearningreally} and majority voting accuracy~\citep{kang2025quagmiressftrlposttraininghigh} may correlate with the RL performance, we show (in Figure~\ref{fig:offline_online}) that the ranking is not always well-preserved when comparing the effectiveness of different techniques either.

\subsection{Why Uniform Loss Is Misaligned}
\label{sec:why_uniform_misalign}

Standard SFT and KL-distillation apply uniform token-level supervision under prefixes induced by the behavior (data-generating) policy \beh, while online RL samples and optimizes rollouts from the evolving target policy \tgt. This behavior–target occupancy mismatch—well known in offline-to-online RL—can hurt the subsequent online phase~\citep{huang2025offlinetoonlinereinforcementlearningclassifierfree,zu2025behavioradaptiveqlearningunifyingframework,lee2021offlinetoonlinereinforcementlearningbalanced,zhao2022adaptivebehaviorcloningregularization}.

\begin{wrapfigure}{r}{0.5\linewidth}
    \centering
    \vspace{-1.0em}
    \includegraphics[width=\linewidth]{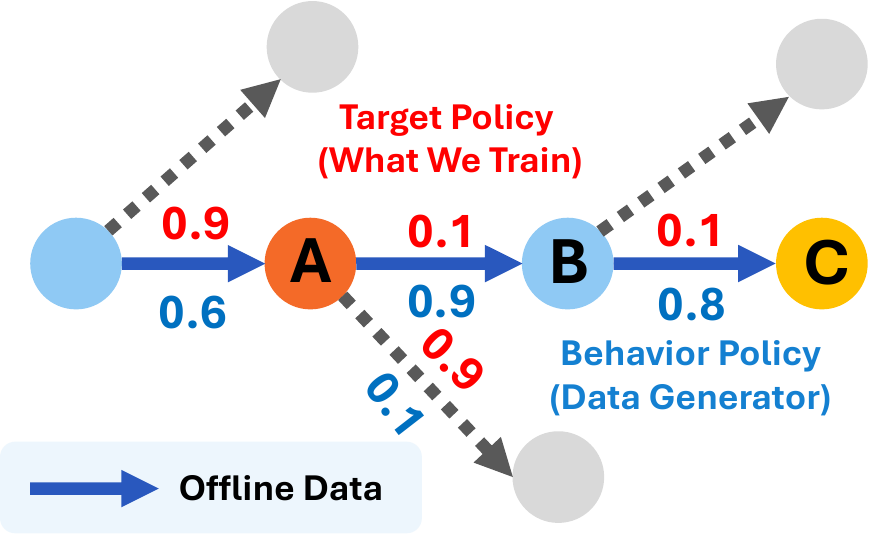}
    \caption{A sketch of our weighing intuition.
    \textbf{\textcolor{red}{Red}} numbers are probabilities under target policy, \textbf{\textcolor{blue}{Blue}} numbers are probabilities under behavior policy.
    After token $A$, the behavior (data-generating) policy often continues with $A\!\rightarrow\!B\!\rightarrow\!C$ (e.g., $0.9\times0.8$), but this continuation is highly unlikely for the policy we ultimately want to optimize.
Therefore, the offline data over-represents $A\!\rightarrow\!B\!\rightarrow\!C$, which can push the model to associate $A$ with an implausible continuation.
During online RL, once the model generates $A$, it will rarely follow with $B$ and $C$, so learning from these offline continuations provides little useful signal.
We thus down-weight token $A$ to avoid visiting it.}
    \label{fig:illustration}
\end{wrapfigure}

In auto-regressive generation, small early mismatches shift the prefix distribution and propagate forward, compounding over long horizons~\citep{ross2011imitation,mehta2024stablebccontrollingcovariateshift,liu2019offpolicypolicygradientstate,ross2014reinforcementimitationlearninginteractive,sun2017deeplyaggrevateddifferentiableimitation}. This is especially acute for long-form reasoning~\citep{Guo_2025}, where traces often involve implicit search (trial, backtracking, self-correction): \beh may over-represent continuations that are effectively dead-ends under \tgt, hence uniformly training on all logged tokens can reinforce transition patterns that RL will rarely revisit for the model, see Fig.~\ref{fig:illustration} for a visualized illustration.
\subsection{Off-Policy Evaluation}
\begin{table}[t]
\scriptsize
\centering
\renewcommand{\arraystretch}{1.15}
\begin{tabular}{c c}
\toprule
\textbf{Name} & \textbf{Per-token objective / weight} \\
\midrule
SFT (NLL) & $\ell(p)=-\log p$ \\
SFT+KL & $\ell(p)=-\log p \;+\; \beta\,\mathrm{KL}$ \\
GeneralFamily-$\alpha$ & $\ell_\alpha(p)=\displaystyle \frac{1-p^{\alpha}}{\alpha}\;\;(\alpha\!\to\!0 \Rightarrow -\log p)$ \\
TopP-$q$ & $(1-p)\,\mathbf{1}[p\ge q]$ \\
BottomP-$q$ & $(1-p)\,\mathbf{1}[p\le q]$ \\
TopLogP-$q$ & $-\log(p)\,\mathbf{1}[p\ge q]$ \\
BottomLogP-$q$ & $-\log(p)\,\mathbf{1}[p\le q]$ \\
TALR & $w_t \propto \exp(-\ell_t/\tau)=p_t^{1/\tau}$ \\
SFT+KL & $-\log(p) + \beta \mathrm{D_{KL}}$ \\
\bottomrule
\end{tabular}
\caption[Compared objectives at the per-token level.]{Compared objectives at the per-token level. Here $p_t:=p_\theta(y_t\mid y_{<t},x)$.}
\label{tab:objective_name_fn}
\end{table}

To reason about the offline-to-online mismatch, we adopt an off-policy evaluation (OPE) lens:
we have logged trajectories from a \emph{behavior} policy $\pi_\beta$, while the subsequent online RL
stage generates rollouts from a (changing) \emph{target} policy $\pi_\theta$.
Classical OPE corrects this behavior--target shift via a change of measure with likelihood ratios
\citep{pdis2000,jiang2016doublyrobustoffpolicyvalue,uehara2022opesurvey,levine2020offlinereinforcementlearningtutorial}:
\(
\mathbb{E}_{\tau\sim\pi_\theta}[f(\tau)]
=
\mathbb{E}_{\tau\sim\pi_\beta}\!\left[\frac{\pi_\theta(\tau)}{\pi_\beta(\tau)}f(\tau)\right].
\)

OPE comprises a family of estimators that correct for behavior–target mismatch using likelihood ratio, including variants that compute likelihood ratio across the entire trajectory $w=\prod{_{t=1}^T} \frac{\pi_{\theta}(a_t|s_t)}{\pi_{\beta}(a_t|s_t)}$~\citep{thomas2016dataefficientoffpolicypolicyevaluation} and uniformly apply to all actions for simplicity~\citep{cis_off_policy} and those that compute suffix ratios from a certain time step $w_n=\prod{_{t=n+1}^T} \frac{\pi_{\theta}(a_t|s_t)}{\pi_{\beta}(a_t|s_t)}$ for each decision~\citep{pdis2000}. This naturally suggests using likelihood-ratio-based sequence or continuation weights as compatibility signals between the logged data and the current policy to correct for the mismatch mentioned in ~\S\ref{sec:why_uniform_misalign} with different granularity.




\section{Method}
\label{sec:method}

To address the offline-to-online mismatch identified in \S~\ref{sec:offline_not_entail_online}, we introduce \name (\fullname), a reweighting scheme for offline fine-tuning that produces a stronger initialization for subsequent online RL on verifiable reasoning. \name keeps the underlying objective unchanged (SFT or KL-based distillation) and modifies only how each token's loss is weighted.

\begin{namesummarybox}

\textbf{Step 1.} Compute token log-likelihood ratios on tokens from offline dataset.\\
\textbf{Step 2.} Aggregate into weights using either one of 3 variants and stabilize it.\\
\textbf{Step 3.} Weigh the loss for each token.

\end{namesummarybox}

\begin{figure}[t]
    \centering
    \includegraphics[width=\linewidth]{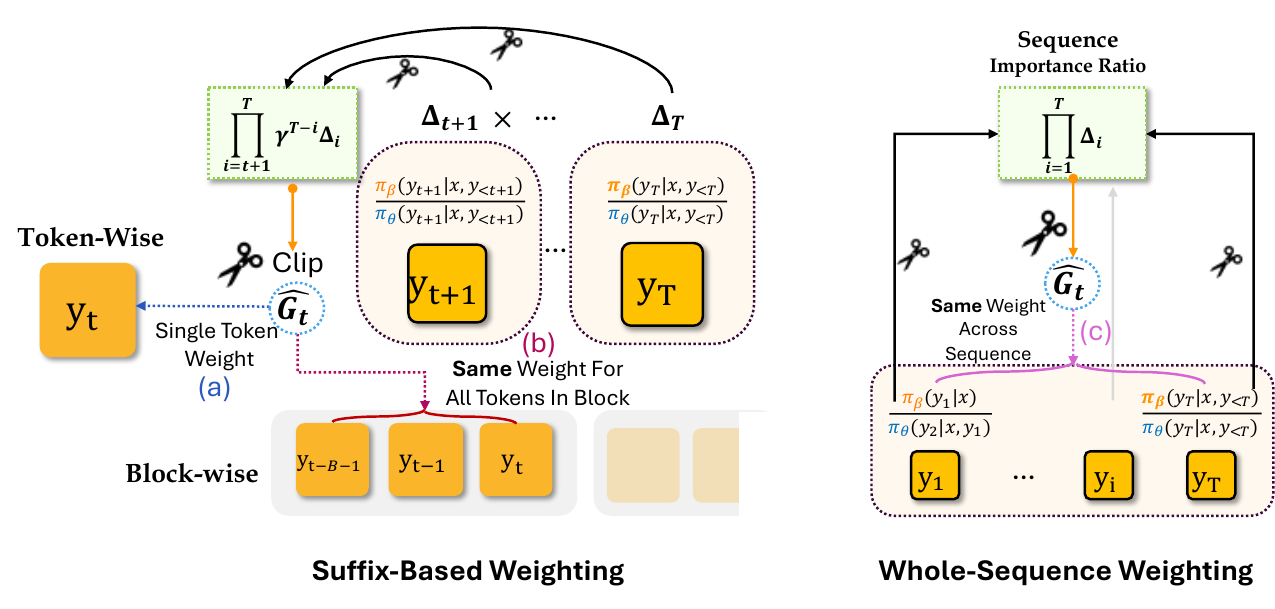}
    \caption{Pipeline of \name. We compute the per-token weight $\widehat{G}_t$ by combining the target-vs.-behavior likelihood ratios over the continuation suffix (change-of-measure following standard OPE), then reweight each token's offline loss accordingly.}
    \label{fig:pipeline}
\end{figure}

\subsection{Problem setup and notation}
\label{sec:setup}

We consider standard offline fine-tuning from a dataset of prompt--response pairs.
$\mathcal{D}=\{(x,\mathbf y)\}$,
where $x$ is a prompt and $\mathbf y$ is a token sequence produced by a known data-generating policy $\pi_\beta$. 
We train a target model $\pi_\theta$ on $\mathcal{D}$.
Let $\pi_\theta(y_t\mid x,\mathbf y_{<t})$ denote the model probability at position $t$, and let
$\ell_\theta(x,\mathbf y_{<t},y_t)$ be a per-token training loss. 

\newcommand{\sg}[1]{\mathop{\mathrm{sg}}\!\left[#1\right]}
\newcommand{\clip}[3]{\mathrm{clip}\!\left(#1,#2,#3\right)}
\newcommand{\StateCont}{\Statex\hspace{\algorithmicindent}}

\begin{algorithm}[h]
\caption{\textsc{\name}}
\label{alg:pear_simple_v2_clean}
\footnotesize
\begin{algorithmic}[1]
\Require One example $(x,y)$ with $y=(y_1,\ldots,y_T)$; model $\pi_\theta$;
\StateCont behavior policy $\pi_\beta$.
\Require Block size $B$; mode $\mathtt{uniform}$ or $\mathtt{suffix}$;
\StateCont discount $\gamma\in(0,1]$.
\Require Token loss $\ell_\theta(x,y_{<t},y_t)$; clip bounds $[\ell_\Delta,u_\Delta]$,
\StateCont $[G_{\min},G_{\max}]$.
\Ensure Weighted loss $L(\theta)$.

\State Partition $\{1,\ldots,T\}$ into $K=\lceil T/B\rceil$ contiguous blocks
      $\{\mathcal I_k\}_{k=1}^K$; \algc{token-level weighting: $B=1$}
\StateCont let $e_k=\max\mathcal I_k$.

\State \textbf{Per-token quantities.}
\State $\delta_t \gets
       \clip{\log\pi_\theta(y_t\mid x,y_{<t})-\log\pi_\beta(y_t\mid x,y_{<t})}
            {\ell_\Delta}{u_\Delta},$
\Statex \hspace{\algorithmicindent} $\forall t\in[T]$
\StateCont \algc{clipped log-ratios for numerical stability}
\State $\ell_t \gets \ell_\theta(x,y_{<t},y_t),\quad \forall t\in[T]$
\StateCont \algc{token loss under the base objective (SFT/KD)}

\State \textbf{Blockwise reductions.}
\State $\rho_k \gets \sum_{t\in\mathcal I_k}\delta_t,\quad \forall k\in[K]$
\StateCont \algc{block log-ratio (log of within-block product of $\pi_\theta/\pi_\beta$)}
\State $b_k \gets \sum_{t\in\mathcal I_k}\ell_t,\quad \forall k\in[K]$
\StateCont \algc{aggregate loss for this block}

\If{$\mathtt{uniform}$}
  \State $\widehat{G}_T \gets \clip{\exp\!\big(\sum_{k=1}^{K}\rho_k\big)}{G_{\min}}{G_{\max}}$
  \State $\widehat{G}_t \gets \widehat{G}_T,\quad \forall t\in[T]$
        
  \State \Return $\displaystyle \sum_{k=1}^{K}\sg{\widehat{G}_T}\, b_k$
        \algc{equivalently $\sum_{t=1}^{T}\sg{\widehat{G}_t}\,\ell_t$}
\EndIf

\State \textbf{Suffix mode: continuation weights (single backward scan).}
\State $L \gets 0$; $u \gets 0$
      \algc{$u$ tracks future log-ratio: $\sum_{m=k+1}^{K}\rho_m$}
\For{$k=K$ downto $1$}
  \State $\widehat{G}_{e_k} \gets \clip{\exp\!\big((T-e_k)\log\gamma + u\big)}{G_{\min}}{G_{\max}}$
        \algc{weight by how plausible the \emph{remaining continuation} is under $\pi_\theta$}
  \State $\widehat{G}_t \gets \widehat{G}_{e_k},\quad \forall t\in\mathcal I_k$
        \algc{same $G_t$ within a block}
  \State $L \gets L + \sg{\widehat{G}_{e_k}}\, b_k$
        
  \State $u \gets u + \rho_k$
\EndFor
\State \Return $L$
\end{algorithmic}
\end{algorithm}

\subsection{\name-weighted Offline Training Objective}
\label{sec:objective}
\name comes as a simple weighting approach for standard offline loss on each token. 
Given an example $(x,\mathbf y)\sim\mathcal D$ from the dataset, we first compute the weight $G_t$ following either one of the 3 weighting strategies we will present below. We will apply numerical stabilization techniques discussed in a subsequent subsection~\S\ref{sec:numerical_stability} and denote the numerically-stabilized version as $\hat{G_t}$. \name keeps the underlying per-token loss $\ell_\theta(x,\mathbf y_{<t},y_t)$ unchanged
(SFT/NLL or KD/forward-KL), and only reweighs it:
\[
\mathcal{L}_\text{\name}(\theta)
\;\triangleq\; 
\mathbb{E}_{(x,\mathbf y)\sim\mathcal D}\!\left[
\sum_{t=1}^{T}
\sg{\widehat G_t}\ell_\theta(x,\mathbf y_{<t},y_t)
\right],
\]
where $\sg{\cdot}$ stops gradients through the weights (we treat $\widehat G_k$ as a
fixed coefficient, not an additional differentiable path).

\subsection{Sequence-Level Weighting}
\label{sec:method_seq_level}

We start by presenting the simplest form of \name weighting: sequence-level importance weighting. For each token, let us define \(\Delta_t
\triangleq\frac{\pi_\theta(y_t\mid x,y_{<t})}{\pi_\beta(y_t\mid x,y_{<t})},\) to denote the probability ratio between policy we want to train $\pi_{\theta}$ and the behavior (data-generating) policy $\pi_{\beta}$. Then, the resulting sequence-level importance ratio is $w_{1:T}
\triangleq
\frac{\pi_\theta(\mathbf y\mid x)}{\pi_\beta(\mathbf y\mid x)}
=
\prod_{t=1}^{T}\Delta_t$ can be used to represent sequence's relative likelihood under \tgt to \beh. This allows us to estimate the loss under the target policy's distribution:
\(
\mathbb{E}_{y\sim \pi_\theta(\cdot\mid x)}\!\big[\ell_{\theta}(x,y)\big]
\;=\;
\mathbb{E}_{y\sim \pi_\beta(\cdot\mid x)}\!\left[
w_{1:T}\,\ \ell_{\theta}(x,y)
\right].
\)
We therefore use this weight $G_{i} \triangleq w_{1:T}\quad \forall i=1,2,..T$ to equally weigh each token in the trajectory. 
In our experiments, we show that this simple weighting mechanism can yield strong performance. 

\subsection{Token-level Weighting Based on Continuation } 
\label{sec:method_token_level}
We then take a more granular view on the sequence. Sequence-level \name uniformly applies the global importance score to each token, yet the weights may not be uniform across positions, a sequence may become `unlikely'' because of implausible regions. 
For a token $y_t$, we evaluate whether the \emph{continuation from dataset} $\mathbf y_{>t}$ remains plausible under the model \emph{conditioned on} taking $y_t$.
If the relative plausibility is small, it means gradients at time $t$ primarily encourage tokens that lead into regions that $\pi_\theta$ is unlikely to revisit when sampling from itself.
We therefore down-weight the loss on such tokens, focusing offline updates on prefixes whose continuations from the dataset are compatible with the current policy. To this effect, we introduce a token-level importance weighting based on the suffix importance ratio, where $G_t = \gamma^{T-t}\prod{_{j=t+1}^T}\Delta_j$ where $\gamma \in(0,1]$ is a discount factor to control variance in long horizon~\citep{sutoonrl,jiang2016doublyrobustoffpolicyvalue}.

\subsection{Block-Level Weighting to Improve Stability}
\label{sec:method_block_level}
A product over long horizons inevitably introduce large variance~\citep{bossens2024lowvarianceoffpolicyevaluation,liu2018breakingcursehorizoninfinitehorizon,liu2020understandingcursehorizonoffpolicy}.
To reduce the effective length of the multiplicative importance-weight for better stability, we present a block-level variant that trades granularity for stability.

We partition positions $\{1,\ldots,T\}$ into $K=\lceil T/B\rceil$ contiguous blocks
$\{\mathcal I_k\}_{k=1}^{K}$, each of length at most $B$.
Let $e_k \triangleq \max \mathcal I_k$ denote the last index of block $k$.

For each block $k$, define the product of token-level ratios within the block as
\(\Delta_k^{\mathrm{blk}} \triangleq \prod_{t\in \mathcal I_k}\Delta_t .\)

Let
\(S_k \triangleq \prod_{j=e_k+1}^{T}\Delta_j\) denote the importance ratio of the suffix after block $k$
(i.e., how likely the remaining continuation is under $\pi_\theta$ relative to $\pi_\beta$).
Equivalently, $S_k$ can be computed block-wise as $S_k=\prod_{m=k+1}^{K}\Delta_m^{\mathrm{blk}}$.

We assign every token in block $k$ the same discounted continuation weight
\(G_t \;=\; G_k^{\mathrm{blk}}
\;\triangleq\;
\gamma^{T-e_k}\,S_k,\forall\, t\in \mathcal I_k.\)

Note that when $B=1$, we recover the token-level \name introduced in~\S\ref{sec:method_token_level}.

\subsection{Optionally Incorporating Negative Examples}
\label{sec:negative_grad}
When $\mathcal D$ contains verified failures, we optionally add a repulsive term that discourages imitating
negative trajectories in a policy-consistent way. Let
$\mathcal{D}^-=\{x,\mathbf y^-\}$ denote failures. We can still compute sequence-level weights and apply a repulsive term on those data points:
\[
\mathcal{L}_{\text{neg}}(\theta)
\triangleq
\mathbb{E}_{(x,\mathbf y^-)\sim \mathcal{D}^-}\!\left[
-\lambda\,\sg{\widehat {G_t^{-}}}
\sum_{t=1}^{T}\ell_\theta(x,\mathbf y^-_{<t},y^-_t)
\right],
\]  where ${\widehat {G_t^{-}}}$ is a sequence level weight on negative trajectories. 
Here, we perform gradient ascent to push the model away from the negative response with a trajectory-level weight.

\subsection{Numerical Stabilization}
\label{sec:numerical_stability}
For numerical stability, we compute importance weights in log-space to avoid products of ratios over long sequences. We apply clip on both per-decision ratios $\Delta_t$ and final weights $\hat{G_t}$ as described in Algorithm~\ref{alg:pear_simple_v2_clean}.
\section{Experiments}
\label{sec:experiments}

\begin{figure*}
\centering
  \begin{subfigure}[t]{0.62\linewidth}
    \centering
    \includegraphics[width=\linewidth]{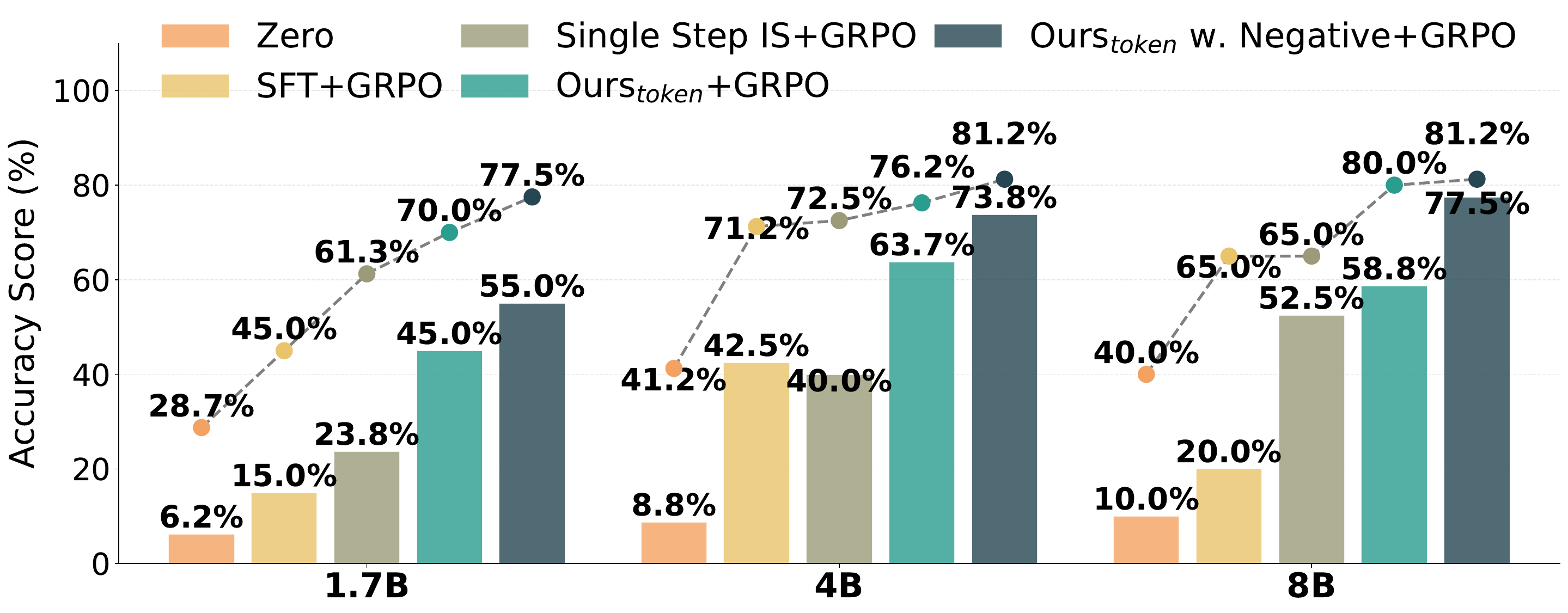}
    \caption{Pass@1 comparison across initializations for SynLogic Games. \namebasic\space significantly improve upon SFT initialization, and incorporating negative gradients can further improve Pass@1.}
    \label{fig:synlogic_results_basic}
  \end{subfigure}
  \hfill
  \begin{subfigure}[t]{0.35\linewidth}
    \centering
    \includegraphics[width=\linewidth]{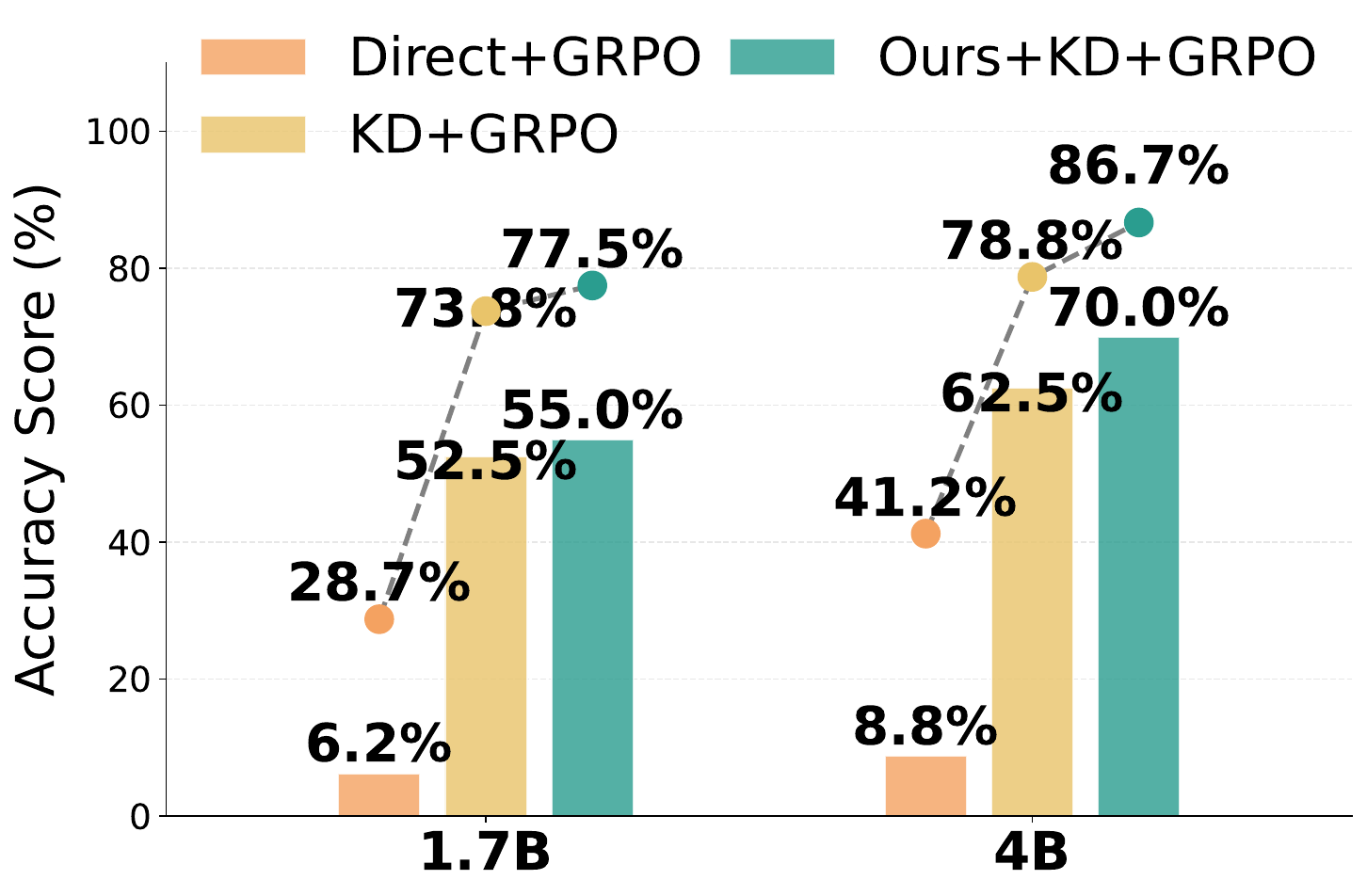}
    \caption{\namebasic applied to KL based knowledge distillation.}
    \label{fig:kd}
  \end{subfigure}
    \caption{Results on SynLogic dataset. We demonstrate that \name consistently improves post-RL performance. The bars reflect Pass@1 and dots mark Pass@8. In Figure~\ref{fig:synlogic_results_basic}, Single Step IS is a baseline that corrects each token only based on the probability ratio of the token itself. See \S\ref{sec:single_action}. }
    \label{fig:synlogic_main_results}
\end{figure*}

We present a careful controlled study under clean set-ups to study the effectiveness of various \name-based weighting, that in turn proves the insights in \S~\ref{sec:offline_not_entail_online}.
Our main experiments target \emph{verifiable reasoning} (math and logic games), where rule-based verifiers eliminate reward confounds; we additionally verify that \name remains effective beyond this regime — on instruction-following tasks (App.~\ref{app:instruction_following}), with proxy/ensemble behavior policies when $\pi_\beta$ is not directly accessible (App.~\ref{app:blackbox}), and paired with DAPO instead of GRPO (App.~\ref{app:pear_dapo}).
\subsection{Tasks and data}
\label{sec:exp_tasks_data}
\subsubsection{Logic games}
\label{sec:exp_games}
\paragraph{Task Sources.}
We use synthetic, verifiable puzzles from SynLogic~\citep{liu2025synlogic} and Enigmata~\citep{chen2025enigmata} as a primary testbed.
Both of them are synthetic reasoning environments that procedurally generate verifiable puzzle instances from diverse environments to allow noise-free data collation, training and evaluation. This allows for a minimally confounded evaluation setting, with reduced exposure to knowledge dependence and contamination.

\paragraph{Offline buffer construction.}
We generate synthetic games using the rule-based generator and de-duplicate prompts across train/test and remove any train instances that overlap with evaluation prompts. We sample responses with Qwen3-8B~\citep{yang2025qwen3technicalreport}
and verify final answers.
The resulting offline buffer contains roughly 100,000 correct trajectories. 


\paragraph{Evaluation.}
We measure Pass@\{1, 8\} on a held-out set of puzzles with the original verifiers. We evaluate using samples from SynLogic's evaluation set.

\subsubsection{Math reasoning}
\label{sec:exp_math}
\paragraph{Data.}
For offline training, we use the subset of all math problems in SYNTHETIC-2 dataset~\citep{primeintellect_synthetic2_2025} -- a total of 33,400 unique instructions. We sample responses from Qwen3-8B and verify with final answer, forming a dataset of 100,000 question-response pairs. For online RL, we use DAPO-17k dataset~\citep{yu2025dapo}.
\paragraph{Evaluation.}
Following common practice, we evaluate on MATH-500~\citep{hendrycks2021MATH}, MINERVA~\citep{lewkowycz2022MINERVA}, AIME-2024, AIME-2025 and AMC-2023~\citep{matharena}. We report average accuracy across 64 samples to reduce variance and pass@K. 

\subsection{Training details}
\label{sec:exp_training_details}

\paragraph{Offline training}
\label{sec:exp_offline_training}
Unless stated otherwise, we train for 1 epoch with learning rate $3\times10^{-5}$ on games and $1\times10^{-5}$ on math.
For \name, we use $\gamma=0.999$, clip $\log\hat{G}_t$ to $[-10,5]$, and clip per-decision $\log\Delta_t$ to $[-0.08,0.3]$.

\begin{table*}[t]
\small
\centering
\renewcommand{\arraystretch}{1.05}

\newcommand{\namecell}[2]{%
  {\color{pearfg}#1\%}\hspace{0.15em}%
  \raisebox{-0.15ex}{\smash{\scriptsize\color{orange}(#2)}}%
}

\begin{tabular}{cc|cP cP cP cP}
\hline
\multicolumn{2}{c|}{\textbf{Base Model}} &
\multicolumn{2}{c}{\textsc{Qwen2.5-1.5B-Math}} &
\multicolumn{2}{c}{\textsc{DS-Qwen-1.5B}} &
\multicolumn{2}{c}{\textsc{Qwen3-4B-Base}} &
\multicolumn{2}{c}{\textsc{Qwen3-8B-Base}} \\
\hline
\textsc{Benchmark} & \textsc{Pass@} &
\shortstack{\textsc{SFT}\\[-0.2ex]{\scriptsize\textsc{+GRPO}}} & \shortstack{{\color{pearfg}\textsc{\namebasic}}\\[-0.2ex]{\scriptsize\color{pearfg}\textsc{+GRPO}}} &
\shortstack{\textsc{SFT}\\[-0.2ex]{\scriptsize\textsc{+GRPO}}} & \shortstack{{\color{pearfg}\textsc{\namebasic}}\\[-0.2ex]{\scriptsize\color{pearfg}\textsc{+GRPO}}} &
\shortstack{\textsc{SFT}\\[-0.2ex]{\scriptsize\textsc{+GRPO}}} & \shortstack{{\color{pearfg}\textsc{\namebasic}}\\[-0.2ex]{\scriptsize\color{pearfg}\textsc{+GRPO}}} &
\shortstack{\textsc{SFT}\\[-0.2ex]{\scriptsize\textsc{+GRPO}}} & \shortstack{{\color{pearfg}\textsc{\namebasic}}\\[-0.2ex]{\scriptsize\color{pearfg}\textsc{+GRPO}}} \\
\hline

\multirow{3}{*}{\textbf{AIME24}} & \textsc{Avg. 64}  &
5\%  & \namecell{8}{+3}  &
1\%  & \namecell{14}{+13} &
13\% & \namecell{15}{+2}  &
15\% & \namecell{19}{+4} \\
& \Pass{8}  &
19\% & \namecell{23}{+4} &
2\%  & \namecell{38}{+36} &
25\% & \namecell{40}{+15} &
35\% & \namecell{41}{+6} \\
& \Pass{64} &
37\% & \namecell{47}{+10} &
7\%  & \namecell{57}{+50} &
43\% & \namecell{60}{+17} &
60\% & \namecell{67}{+7} \\
\hdashline

\multirow{3}{*}{\textbf{AIME25}} & \textsc{Avg. 64}  &
3\%  & \namecell{8}{+5}  &
1\%  & \namecell{14}{+13} &
7\%  & \namecell{15}{+8}  &
17\% & \namecell{18}{+1} \\
& \Pass{8}  &
14\% & \namecell{24}{+10} &
5\%  & \namecell{35}{+30} &
21\% & \namecell{35}{+14} &
35\% & \namecell{35}{0} \\
& \Pass{64} &
30\% & \namecell{40}{+10} &
17\% & \namecell{53}{+36} &
40\% & \namecell{57}{+17} &
53\% & \namecell{53}{0} \\
\hdashline

\multirow{3}{*}{\textbf{AMC23}} & \textsc{Avg. 64}  &
42\% & \namecell{47}{+5}  &
20\% & \namecell{56}{+36} &
54\% & \namecell{59}{+5}  &
45\% & \namecell{55}{+10} \\
& \Pass{8}  &
80\% & \namecell{78}{-2}  &
50\% & \namecell{91}{+41} &
83\% & \namecell{88}{+5}  &
75\% & \namecell{85}{+10} \\
& \Pass{64} &
98\% & \namecell{98}{0} &
63\% & \namecell{98}{+35} &
93\% & \namecell{95}{+2}  &
95\% & \namecell{98}{+3} \\
\hdashline

\multirow{2}{*}{\textbf{Olympiad}} & \textsc{Avg. 64} &
29\% & \namecell{34}{+5}  &
13\% & \namecell{41}{+28} &
41\% & \namecell{46}{+5}  &
32\% & \namecell{40}{+8} \\
& \Pass{8} &
55\% & \namecell{57}{+2}  &
34\% & \namecell{66}{+32} &
60\% & \namecell{66}{+6}  &
60\% & \namecell{65}{+5} \\
\hdashline

\multirow{2}{*}{\textbf{MATH500}} & \textsc{Avg. 64} &
63\% & \namecell{70}{+7}  &
36\% & \namecell{72}{+36} &
78\% & \namecell{80}{+2}  &
66\% & \namecell{74}{+8} \\
& \Pass{8} &
89\% & \namecell{91}{+2}  &
68\% & \namecell{94}{+26} &
93\% & \namecell{93}{0}  &
90\% & \namecell{93}{+3} \\
\hline

\multicolumn{2}{c|}{\textbf{Pass@1 Avg.}} &
28\% & \namecell{33}{+5} &
14\% & \namecell{39}{+25} &
39\% & \namecell{43}{+4} &
35\% & \namecell{41}{+6} \\
\multicolumn{2}{c|}{\textbf{Pass@8 Avg.}} &
51\% & \namecell{55}{+4} &
32\% & \namecell{65}{+33} &
56\% & \namecell{65}{+9} &
59\% & \namecell{64}{+5} \\
\hline
\end{tabular}

\caption{\textbf{\textcolor{purple}{After-GRPO}} math results with models initialized by standard SFT v.s. vanilla \name-weighted NLL. The \textsc{Avg.\ 64} rows report mean correctness over 64 sampled responses per question (a lower-variance estimate of per-sample Pass@1); single-sample Pass@1 can be noisy on small benchmarks, so averaging stabilizes the comparison. The \textbf{Pass@1 Avg.}\ row at the bottom averages these per-benchmark \textsc{Avg.\ 64} values across benchmarks. The \name-over-SFT gains are largest on \textsc{DS-Qwen-1.5B}: this model is a product of cross-family distillation, which induces a larger behavior--target divergence---precisely the regime \name is designed to correct, so gains scale with mismatch.
}
\label{tab:math_results}
\end{table*}


\paragraph{Online RL}
\label{sec:exp_online_training}
Starting from each offline checkpoint $\pi_0$, we run the same online RL procedure to obtain $\pi_{\mathrm{RL}}$.
We use GRPO~\citep{shao2024deepseekmath} with learning rate $10^{-6}$, batch size 128, and KL coefficient 0.01.

\subsection{Results}
\label{sec:results}
\namebasic\space stands for token-level weighting (\S\ref{sec:method_token_level}), our default form of \name in the evaluation below that directly reflects the key intuition.
\paragraph{\name improves post-RL performance under a fixed RL budget.}
Figure~\ref{fig:synlogic_results_basic} demonstrates the gain from initializing the model with the token-wise form of \name over standard SFT across different model sizes. We show a clear improvement on pass@1 across model sizes. Moreover, \name outperforms all previously mentioned techniques in \S~\ref{sec:offline_not_entail_online}. Notably, \name's performance does not surface in terms of its \emph{out-of-the-box offline performance}, and even may not beat SFT, since \name is not designed to boost offline scores in isolation, but to better shape the prior for online RL. 

In addition, we show in Table~\ref{tab:math_results} that \name-initialized model can achieve higher overall performance across common math reasoning benchmarks for multiple model families. 
\newcommand{\sftcell}[1]{%
  \shortstack[c]{#1\%\\[-0.25ex]\scriptsize\phantom{(+0)}}%
}

\newcommand{\gaincellc}[4]{%
  \shortstack[c]{\textcolor{#1}{#2\%}\\[-0.25ex]\scriptsize\textcolor{#4}{(#3)}}%
}

\newcommand{\gainposc}[3]{\gaincellc{#1}{#2}{#3}{orange}} 
\newcommand{\gainnegc}[3]{\gaincellc{#1}{#2}{#3}{red}}    

\begin{table}[t]
\centering
\small
\renewcommand{\arraystretch}{1.05}
\setlength{\tabcolsep}{4pt}
\begin{tabular}{@{} lcccc @{}}
\hline
\textsc{Qwen3-Base} & \textsc{0.6B} & \textsc{1.7B} & \textsc{4B} & \textsc{8B} \\
\hline
\shortstack[c]{\textsc{SFT}\\[-0.2ex]{\footnotesize\textsc{+GRPO}}}
  & \sftcell{9}  & \sftcell{23} & \sftcell{39} & \sftcell{36} \\
\hdashline
\shortstack[c]{\textsc{Single Step}\\[-0.2ex]{\footnotesize\textsc{+GRPO}}}
  & \gainposc{black}{10}{+1} & \gainnegc{black}{18}{-5} & \gainnegc{black}{38}{-1} & \gainnegc{black}{30}{-6} \\
\hdashline
\shortstack[c]{{\color{pearfg}\textsc{\namebasic}}\\[-0.2ex]{\footnotesize\color{pearfg}\textsc{+GRPO}}}
  & \gainposc{pearfg}{12}{+3} & \gainposc{pearfg}{26}{+3} & \gainposc{pearfg}{44}{+5} & \gainposc{pearfg}{41}{+5} \\
\hline
\end{tabular}
\caption{Average accuracy across math benchmarks: parentheses denote changes over \textsc{SFT+GRPO} (\%). Negative gains are shown in red.}
\label{tab:math-aggregate}
\end{table}

Figure~\ref{fig:kd} shows that \name's weighting scheme can also work with KL-based knowledge distillation ($\ell_\theta=\mathrm{KL}(\pi_\beta(\cdot\mid x,y_{<t})\,\|\,\pi_\theta(\cdot\mid x,y_{<t}))$) and further improve upon that by computing the score using already-computed information during KD, adding minimal overhead to the KL-based KD baseline. \namebasic\space uses the exact same weighting as we apply it with NLL loss (i.e. suffix likelihood ratio).

This underscores \name’s generality as a plug-in reweighting approach for both commonly used offline objectives. Importantly, it proves our central hypothesis that offline stage should correct for distribution mismatch between behavior and target policies.

\begin{figure}
    \centering
\includegraphics[width=.65\linewidth]{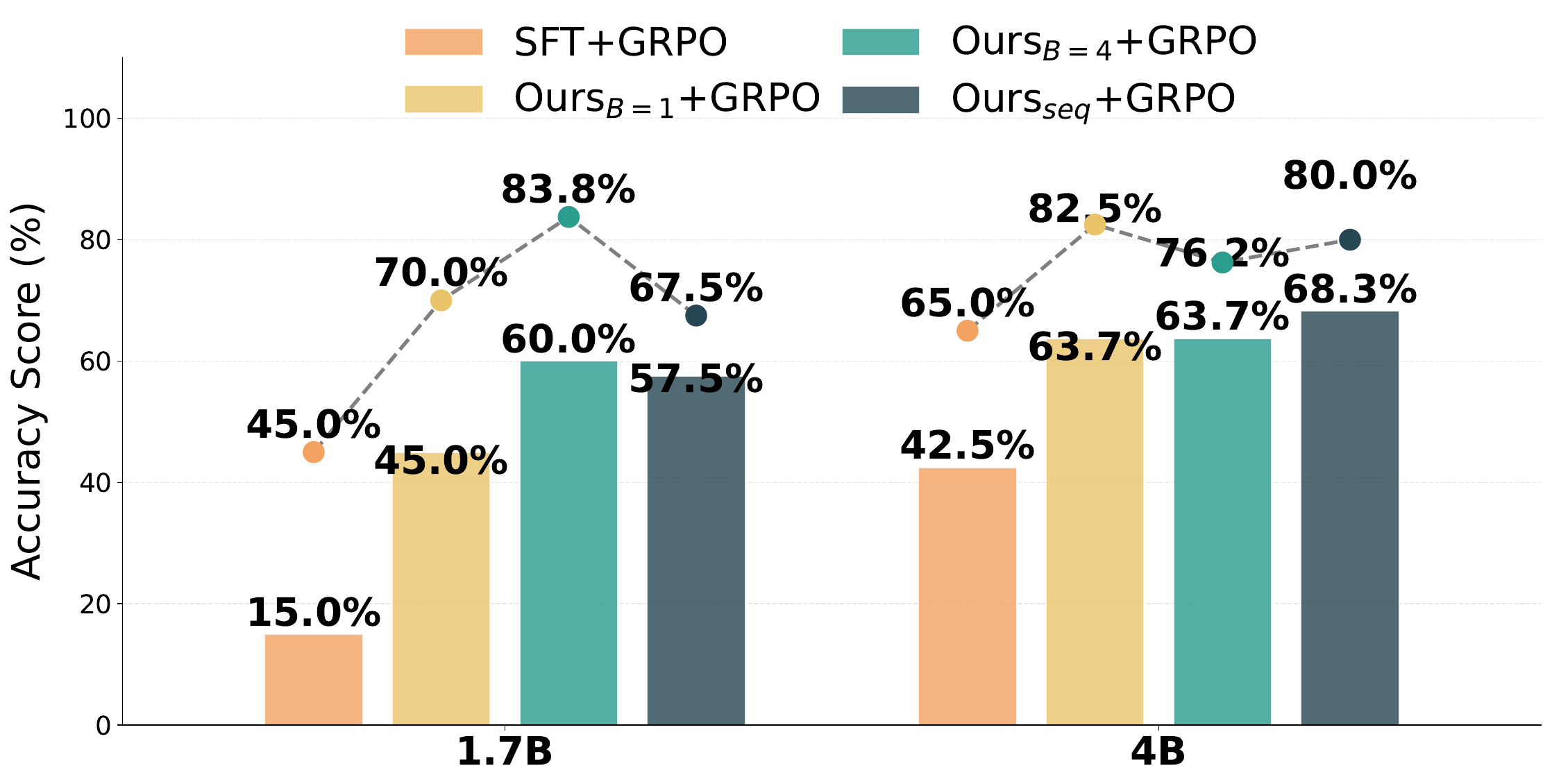}
    \caption{Performance of different variants of \name. The bars reflect Pass@1 and dots mark Pass@8. 
    }
    \label{fig:synlogic_blocksize}
    \vspace{-1mm}
\end{figure}

In Figure~\ref{fig:synlogic_blocksize}, we compare different modes of \name: Sequence-Level(\S\ref{sec:method_seq_level}), Token-Level(\S\ref{sec:method_token_level}) and Block-Level(\S\ref{sec:method_block_level}). All these variants out-perform standard SFT. Additionally, we observe that sequence-level weighing turns out highly effective despite its simplicity. 

\paragraph{You need to weigh the future, not a single action.}
\label{sec:single_action}
Concurrent works~\citep{wu2025dft,zhang2025onpolicyrlmeetsoffpolicy,zhu2025proximalsft,zhu2025aft}  propose several action-level stabilization to SFT of the form
\[
\mathcal{L}(\theta)
\;=\;
\mathbb{E}_{(x,y,R)\sim\mathcal{D}}
\left[
\sum_{t=1}^{T} {\color{purple}\mathbf{w(x,y_{<t},y_t)}}\ell_\theta(x,y_{<t},y_t)
\right],\] with $\mathbf{\textcolor{purple}{w}}$ depending only on the prefix and the current action. We experiment with $w(x,y_{<t},y_t) = \frac{\pi_\theta(y_t|x,y{<t})}{\pi_\beta(y_t|x,y{<t})}$ which is a generic form of one-step weighting, computed and stabilized the same way as $\Delta_t$ in \name. 

\begin{figure}[t]
    \centering
    \includegraphics[width=.62\linewidth]{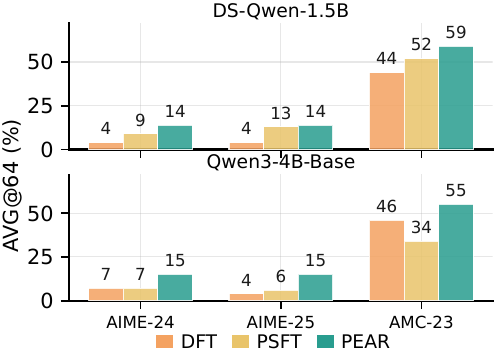}
    \caption{Comparison with importance-sampling baselines DFT~\citep{wu2025dft} and Proximal-SFT~\citep{zhu2025proximalsft} on math reasoning (AVG@64). \name consistently outperforms both on AIME-24, AIME-25, and AMC-23.}
    \label{fig:dft_psft_main}
\end{figure}

\begin{figure}[t]
    \centering
    \includegraphics[width=.62\linewidth]{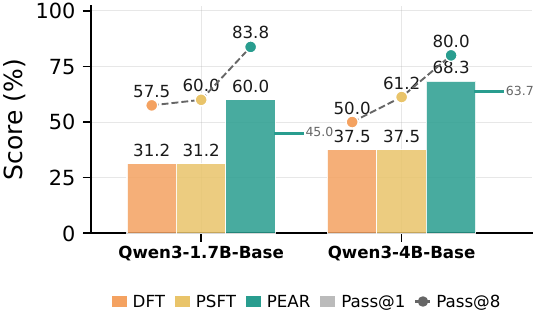}
    \caption{Same DFT/PSFT comparison on SynLogic. Bars = Pass@1 (best \name config), dots + dashed line = Pass@8. The small right-side tick on each \name bar marks \name's worst-config Pass@1 — even the worst \name run substantially beats both baselines.}
    \label{fig:synlogic_comparison_main}
\end{figure}

This is a myopic objective that up-weights single actions that the target policy finds plausible, not taking into account the long-term effect. As shown in Figure~\ref{fig:synlogic_results_basic} and Table~\ref{tab:math-aggregate}, single-step weighting is less effective, since what matters for online RL readiness is whether the logged successful continuation is compatible with the current policy over the remaining horizon~\citep{jiang2016doublyrobustoffpolicyvalue,metelli2020ispo,nachum2019dualdicebehavioragnosticestimationdiscounted,ross2011imitation}.We further compare \name against two concurrent importance-sampling-style baselines, DFT~\citep{wu2025dft} and Proximal-SFT~\citep{zhu2025proximalsft}, on math reasoning (Figure~\ref{fig:dft_psft_main}) and SynLogic (Figure~\ref{fig:synlogic_comparison_main}). \name dominates both across model families and benchmarks.

\newcommand{\gaincell}[2]{%
  {\color{pearfg}#1\%}\hspace{0.15em}%
  \raisebox{-0.15ex}{\smash{\scriptsize\color{orange}(#2)}}%
}

\begin{table}[t]
\centering
\small
\renewcommand{\arraystretch}{1.05}
\begin{tabular}{@{} lccc @{}}
\hline
 & \shortstack{\textsc{Direct}\\[-0.2ex]{\footnotesize\textsc{+GRPO}}}
 & \shortstack{\textsc{SFT}\\[-0.2ex]{\footnotesize\textsc{+GRPO}}}
 & \shortstack{{\color{pearfg}\textsc{\namebasic}}\\[-0.2ex]{\footnotesize\color{pearfg}\textsc{+GRPO}}} \\ \hline
\textsc{Qwen3-0.6B-Base} & 2.8\%  & 8.4\%  & \gaincell{13.1}{+4.7} \\
\textsc{Qwen3-1.7B-Base} & 2.8\%  & 13.1\% & \gaincell{38.3}{+25.2} \\
\textsc{Qwen3-4B-Base}   & 8.0\%  & 49.5\% & \gaincell{59.8}{+10.3} \\
\textsc{Qwen3-8B-Base}   & 15.0\% & 53.3\% & \gaincell{61.7}{+8.4} \\ \hline
\end{tabular}
\caption{\name can transfer to different RL task distribution.}
\label{tab:synlogic_enigmata}
\end{table}
\subsection{\name Transfers to different RL task distributions.}
\label{sec:transfer}
\begin{figure}
\centering
  \begin{subfigure}[t]{0.38\linewidth}
    \centering
    \includegraphics[width=\linewidth]{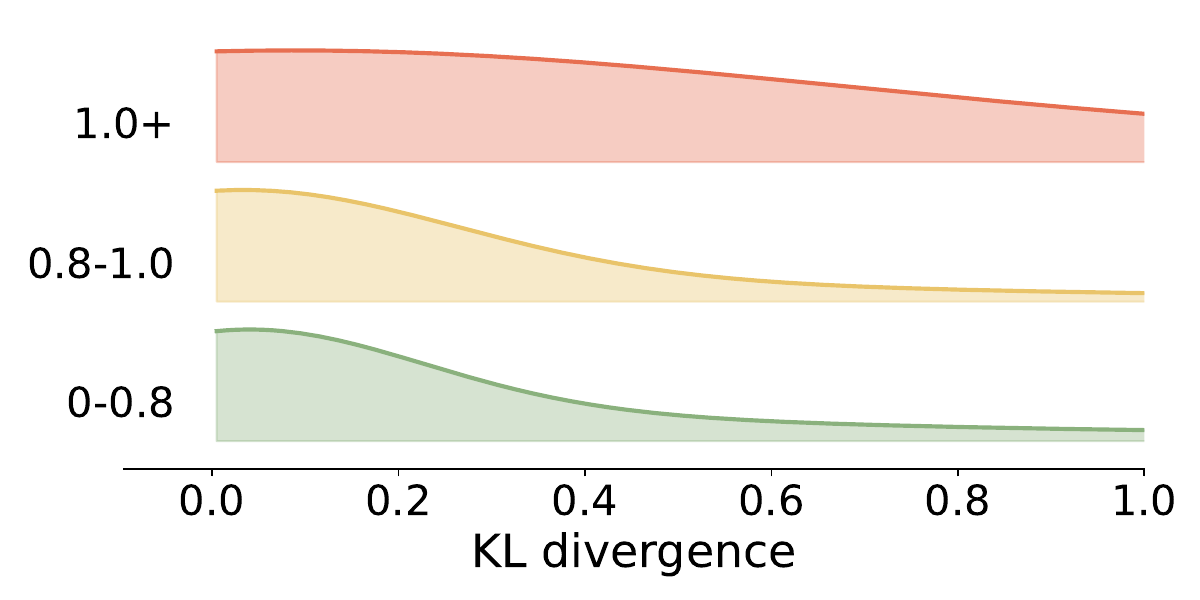}
    \caption{Qwen3-1.7B-Base}
    \label{fig:kl_1_7b}
  \end{subfigure}
  \begin{subfigure}[t]{0.38\linewidth}
    \centering
    \includegraphics[width=\linewidth]{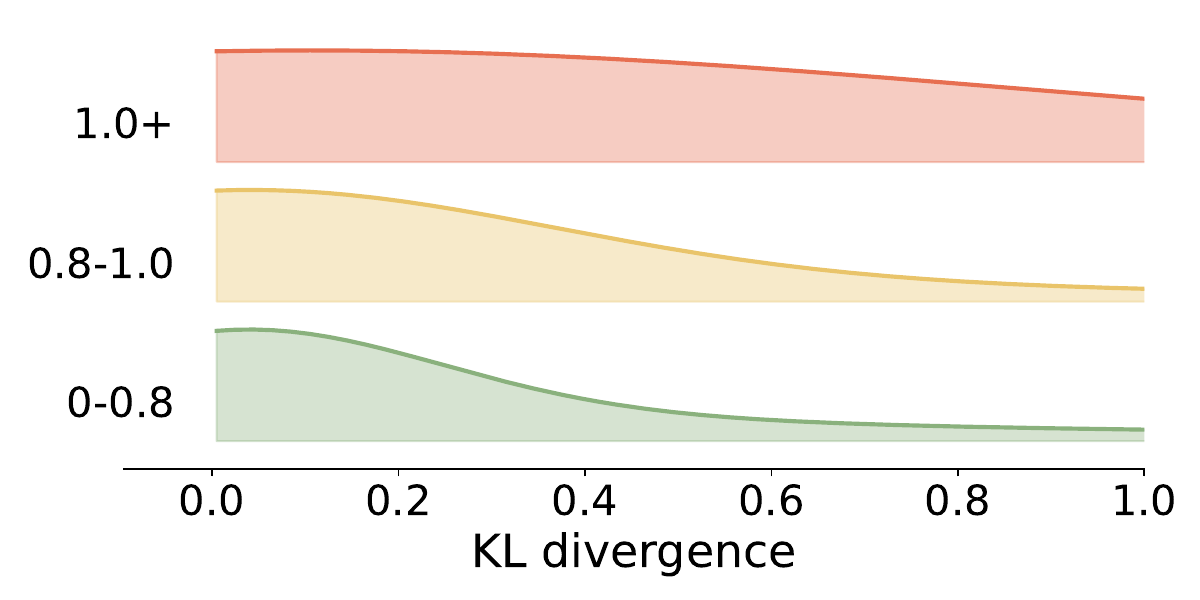}
    \caption{Qwen3-4B-Base}
    \label{fig:kl_4b}
  \end{subfigure}
    \caption{\name-to-base KL divergence across weight levels. $y$-axis is the weight (clipped). The token distribution is more heavily steered on important tokens that drive success probability.}
    \label{fig:kl}
\end{figure}

We next test whether the capability induced by \name during offline training transfers to online RL on a
different task distribution. Concretely, we initialize RL from the \name checkpoint and run online training
on a subset of 12.8K problems from the Enigmata training set, then evaluate on a held-out set of
Enigmata tasks after removing any near-duplicates across splits.
As shown in
Table~\ref{tab:synlogic_enigmata}, despite the difference between the offline training domain and the
online RL domain, \name consistently provides a stronger initialization than standard SFT: it achieves better post-RL performance under the same RL recipe and roll-out budget.
The benefit of \name is not overfit to the offline domain; it transfers better to a shifted online RL distribution under identical RL compute.

\subsection{Black-Box \name: Proxy Behavior Policies}
\label{sec:blackbox}
\begin{figure}[t]
    \centering
    \includegraphics[width=.58\linewidth]{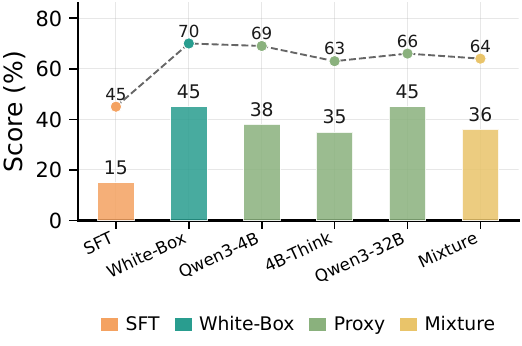}
    \caption{Black-box \name on SynLogic (Qwen3-1.7B-Base). Bars = Pass@1; markers = Pass@8. \textsc{White-Box} uses the true behavior policy $\pi_\beta$; remaining \name variants use an approximate proxy. All proxy variants substantially outperform SFT, and the strongest proxy (Qwen3-32B) matches or exceeds white-box. Full results including Qwen3-4B-Base are in App.~\ref{app:blackbox}.}
    \label{fig:blackbox_pear_main}
\end{figure}
\name requires token log-probabilities from $\pi_\beta$ to compute importance weights. This is straightforward when the offline data is generated by a known open-weight model (as in our main experiments), but may be impractical when SFT data is curated or produced by a closed-source teacher. We show that \name remains effective when $\pi_\beta$ is approximated by a \emph{proxy} model — a different Qwen3 variant not used to generate the data — or an \emph{ensemble} of proxies.

As shown in Figure~\ref{fig:blackbox_pear_main}, every proxy variant substantially outperforms SFT+GRPO.
The strongest proxy (Qwen3-32B) matches white-box \name on Pass@1 and comes within a few points on Pass@8.
This demonstrates that \name does not require exact $\pi_\beta$ access: a reasonably aligned proxy suffices.

\subsection{Incorporating Negative Examples}
\label{sec:exp_negative_grad}
We experiment with the variant presented in~\S\ref{sec:negative_grad}, which pushes down the likelihood of entire
negative sequences while avoiding token-/suffix-level signed ratio products that can be particularly unstable on long horizons. We sub-sampled 50K positive data and included 50K negative data from the same behavior policy on the same set of instructions. 

Figure~\ref{fig:synlogic_results_basic} shows that under the same offline data budget, mixing negative trajectories for stabilization can bring significant additional gains to RL over positive-only \name initialization.

\subsection{Analysis}
\label{sec:analysis}
\paragraph{RQ1: What Positions Does \name Concentrate On?}

By design, \name's learning concentrates on the tokens that evaluate to larger weights. To analyze this effect, we compute per-token weight (\S~\ref{sec:method_token_level}) $\hat{G_t}$'s and $\text{KL}(\pi_\theta(a_t|s_t) \,\|\, \pi_\beta(a_t|s_t))$. As shown in Figure~\ref{fig:kl}, the high-value tokens are distributionally more steered away from the base policy $\pi_\beta$, showing that the behavior of the trained model is systematically more updated on those important locations steering the suffix. 

\begin{wrapfigure}{l}{0.35\textwidth}
  \centering
  \includegraphics[width=\linewidth]{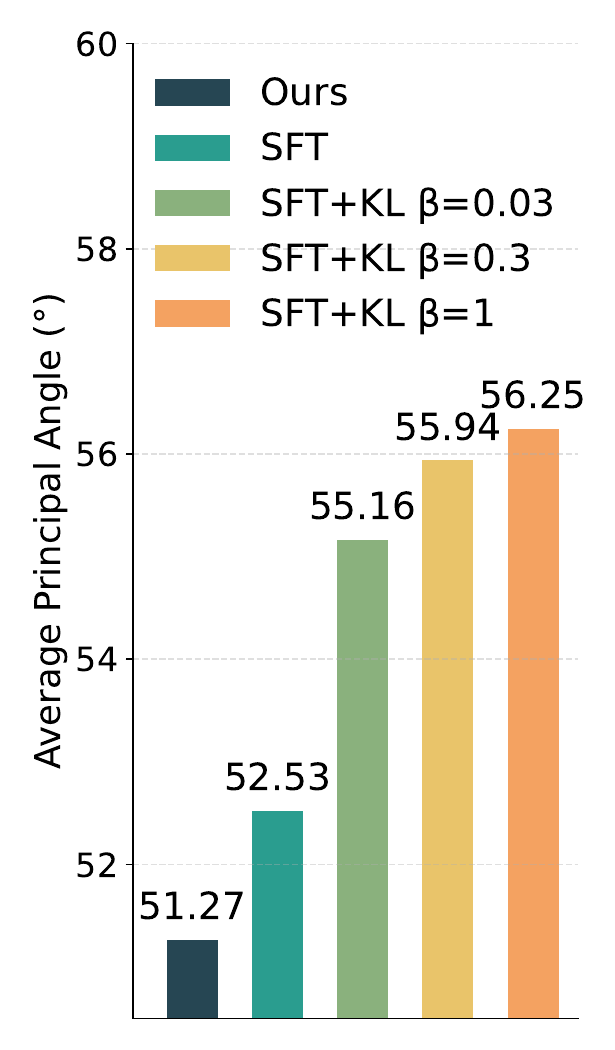}
  \caption{Mean principal angle between offline and online GRPO gradients.}
  \label{fig:rotation_angle}
  \vspace{-1\baselineskip} 
\end{wrapfigure}

\paragraph{RQ2: How Does \name's Learning Interact with Online RL?} 
Figure~\ref{fig:rotation_angle} shows the average principal angle between \name's gradients and those for GRPO is smaller than those of SFT and variants, suggesting that \name's correction can indeed make the offline updates more consistent with the online GRPO learning direction.

We also observe that applying stronger KL constraints could create greater mismatch between offline and online gradients, although it better preserves closeness to base model.

\begin{figure}
\centering
\includegraphics[width=.5\linewidth]{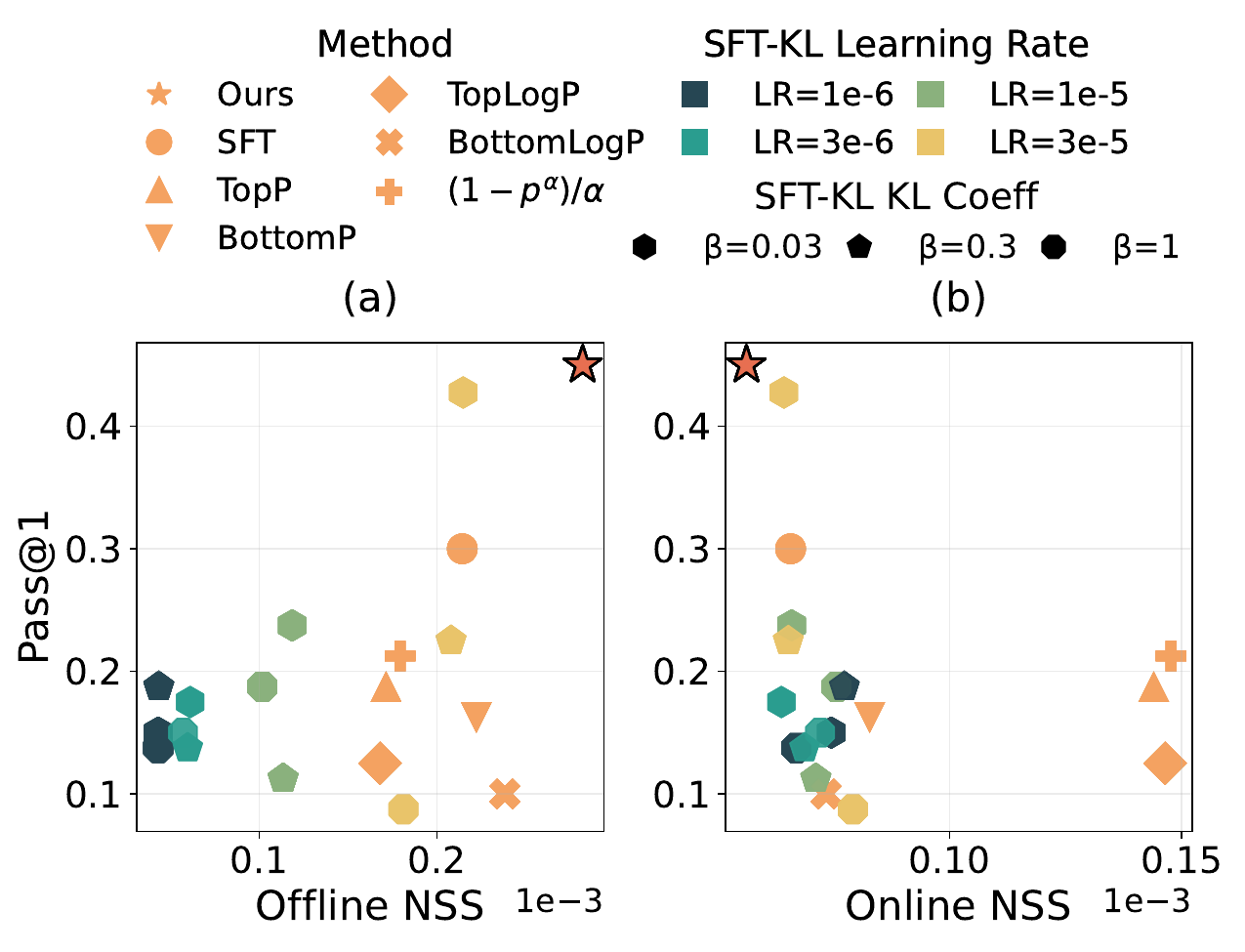}
    \caption{Parameter drift of different approaches. (a) is the NSS score between offline and base model. (b) is the NSS score between online and offline model.
    }
\label{fig:model_characteristics}
\vspace{-1mm}
\end{figure}

Thus, we observe online RL training after \name smallest drift measured by average NSS\footnote{NSS measures the relative drift of a singular-value spectrum after training~\citep{zhu2025pathnottaken}} in Figure~\ref{fig:model_characteristics}-b between online and offline checkpoints compared with other initializations, whereas the heavy-lifting happened in the offline stage (Figure~\ref{fig:model_characteristics}-a). It shows \name suffers the least from offline-to-online mismatch and spent less parameter updates correcting for those mis-alignments. 

\section{Related Works}
\paragraph{Learning Dynamics of Post-Training}
There is a growing interest in understanding the learning characteristics of different post-training approaches (SFT v.s. RL).

A growing line of work studies why reinforcement-learning (RL) can behave qualitatively
differently from supervised fine-tuning (SFT).
They find that SFT more readily overfits and degrades out-of-distribution (OOD)
performance, whereas on-policy RL more often improves generalization across distribution shifts and can
partially undo SFT-induced drift \citep{chu2025sftmemorizes,jin2025rlpanacea}.

Recent analyses further connect RL's reduced catastrophic forgetting to its on-policy sampling bias,\citep{shenfeld2025rlrazor,chen2025retaining,jin2025rlfinetuninghealsood}.
Beyond behavior-level metrics, recent analyses probe \emph{parameter-space} dynamics:~\citet{zhu2025pathnottaken} characterize RLVR updates as structured “off-principal” learning that preserves spectral structure relative to SFT, while~\citet{zhao2025echochamber} show RL post-training can amplify patterns already present in pretraining, often concentrating probability mass onto a dominant output mode. 

\paragraph{Offline RL For Language Models} There is a line of work in LM post training that treat responses as logged decision trajectories~\citep{lanchantin2025bridgingofflineonlinereinforcement,wang2024oreo,snell2023offlinerlnaturallanguage,baheti2024leftoverlunchadvantagebasedoffline,richemond2024offlineregularisedreinforcementlearning,mukherjee2025offlinerlrewardweightedfinetuning} and apply policy-optimization techniques to improve performance. 
Others seek to improve online RL by introducing offline / semi-offline mechanisms~\citep{lanchantin2025bridgingofflineonlinereinforcement,zhang2025beyondonline,li2025reporeplayenhancedpolicyoptimization}.
Differently, our focus is on better bridging offline and offline stages in common SFT+RL post-training pipeline. 



\paragraph{Modifications To SFT}

Some works modify the SFT loss itself to reduce overfitting and capability loss—e.g., probability-based objectives beyond NLL~\citep{li2025beyond}, entropy-regularized distribution matching~\citep{diao2026entropy}, and token/sample-wise reweighting or gating to suppress destructive gradients.~\citep{sanyal2025upweightingeasysamplesfinetuning,lin2025sftdoesnthurtgeneral}.There are also various ``importance-weighted / stabilized SFT'' methods like iw-SFT~\citep{qin2025iwsft}, DFT~\citep{wu2025dft}, AFT~\citep{zhu2025aft}, Proximal-SFT~\citep{zhu2025proximalsft}, OPC-SFT~\citep{zhang2025onpolicyrlmeetsoffpolicy} that use probability-ratio or trust-region style weights primarily to mitigate off-policy instability, suppress the influence of low-probability tokens, and constrain KL/entropy drift so that supervised fine-tuning remains well-behaved under distribution shift. In contrast, our approach is not introduced as a stabilization or trust-region heuristic for SFT but a mechanism to better initialize models for subsequent online RL.  


\section{Conclusion}

Reasoning LLM post-training typically follows an offline SFT $\rightarrow$ online RL pipeline, so offline objectives should be judged by how well they initialize RL, not just by SFT accuracy. Across extensive experiments, we find that stronger offline performance is an unreliable proxy for post-RL performance: objectives that dominate after SFT can be overtaken after identical RL, producing substantial rank reversals.

We attribute this gap to offline-to-online policy mismatch. Offline SFT imitates logged continuations from logged prefixes, whereas online RL updates the model on trajectories sampled from its current policy, concentrating learning on prefixes it actually reaches. To reduce this mismatch, we propose \name (\fullname), an OPE-inspired loss reweighting scheme that down-weights logged continuations that are implausible under the current policy and up-weights those that remain plausible. Empirically, \name consistently improves post-RL accuracy across verifiable reasoning games and math benchmarks, yielding up to $30$ percentage points Pass@8 gain on AIME-2025 after online RL. More broadly, our results suggest a practical principle: the offline stage should prioritize \emph{reproducible successes} under the target policy that will be optimized online.

\section*{Acknowledgements}
This work is supported by NSF Grant No.\ CHE2505932, an Amazon AICE Award, gift funding from AI2, and a grant from Coefficient Giving.

\small
\bibliography{references}

\begin{thebibliography}{62}
\providecommand{\natexlab}[1]{#1}
\providecommand{\url}[1]{\texttt{#1}}
\expandafter\ifx\csname urlstyle\endcsname\relax
  \providecommand{\doi}[1]{doi: #1}\else
  \providecommand{\doi}{doi: \begingroup \urlstyle{rm}\Url}\fi

\bibitem[Baheti et~al.(2024)Baheti, Lu, Brahman, Bras, Sap, and
  Riedl]{baheti2024leftoverlunchadvantagebasedoffline}
Ashutosh Baheti, Ximing Lu, Faeze Brahman, Ronan~Le Bras, Maarten Sap, and Mark
  Riedl.
\newblock Leftover lunch: Advantage-based offline reinforcement learning for
  language models, 2024.
\newblock URL \url{https://arxiv.org/abs/2305.14718}.

\bibitem[Balunović et~al.(2026)Balunović, Dekoninck, Petrov, Jovanović, and
  Vechev]{matharena}
Mislav Balunović, Jasper Dekoninck, Ivo Petrov, Nikola Jovanović, and Martin
  Vechev.
\newblock Matharena: Evaluating llms on uncontaminated math competitions, 2026.
\newblock URL \url{https://arxiv.org/abs/2505.23281}.

\bibitem[Bossens and Thomas(2024)]{bossens2024lowvarianceoffpolicyevaluation}
David~M. Bossens and Philip~S. Thomas.
\newblock Low variance off-policy evaluation with state-based importance
  sampling, 2024.
\newblock URL \url{https://arxiv.org/abs/2212.03932}.

\bibitem[Chen et~al.(2025{\natexlab{a}})Chen, Razin, Narasimhan, and
  Chen]{chen2025retaining}
Howard Chen, Noam Razin, Karthik Narasimhan, and Danqi Chen.
\newblock Retaining by doing: The role of on-policy data in mitigating
  forgetting, 2025{\natexlab{a}}.
\newblock URL \url{https://arxiv.org/abs/2510.18874}.

\bibitem[Chen et~al.(2025{\natexlab{b}})Chen, He, Yuan, Chen, Cai, Dai, Yu, Yu,
  Li, Chen, Zhou, and Wang]{chen2025enigmata}
Jiangjie Chen, Qianyu He, Siyu Yuan, Aili Chen, Zhicheng Cai, Weinan Dai,
  Hongli Yu, Qiying Yu, Xuefeng Li, Jiaze Chen, Hao Zhou, and Mingxuan Wang.
\newblock Enigmata: Scaling logical reasoning in large language models with
  synthetic verifiable puzzles, 2025{\natexlab{b}}.
\newblock URL \url{https://arxiv.org/abs/2505.19914}.

\bibitem[Chu et~al.(2025)Chu, Zhai, Yang, Tong, Xie, Schuurmans, Le, Levine,
  and Ma]{chu2025sftmemorizes}
Tianzhe Chu, Yuexiang Zhai, Jihan Yang, Shengbang Tong, Saining Xie, Dale
  Schuurmans, Quoc~V. Le, Sergey Levine, and Yi~Ma.
\newblock Sft memorizes, rl generalizes: A comparative study of foundation
  model post-training, 2025.
\newblock URL \url{https://arxiv.org/abs/2501.17161}.

\bibitem[Diao et~al.(2026)Diao, Yang, Gong, Zhang, Yan, Han, Liang, Xu, and
  Ma]{diao2026entropy}
Muxi Diao, Lele Yang, Wuxuan Gong, Yutong Zhang, Zhonghao Yan, Yufei Han,
  Kongming Liang, Weiran Xu, and Zhanyu Ma.
\newblock Entropy-adaptive fine-tuning: Resolving confident conflicts to
  mitigate forgetting, 2026.
\newblock URL \url{https://arxiv.org/abs/2601.02151}.

\bibitem[Guo et~al.(2025)Guo, Yang, Zhang, Song, Wang, Zhu, Xu, Zhang, Ma, Bi,
  Zhang, Yu, Wu, Wu, Gou, Shao, Li, Gao, Liu, Xue, Wang, Wu, Feng, Lu, Zhao,
  Deng, Ruan, Dai, Chen, Ji, Li, Lin, Dai, Luo, Hao, Chen, Li, Zhang, Xu, Ding,
  Gao, Qu, Li, Guo, Li, Chen, Yuan, Tu, Qiu, Li, Cai, Ni, Liang, Chen, Dong,
  Hu, You, Gao, Guan, Huang, Yu, Wang, Zhang, Zhao, Wang, Zhang, Xu, Xia,
  Zhang, Zhang, Tang, Zhou, Li, Wang, Li, Tian, Huang, Zhang, Wang, Chen, Du,
  Ge, Zhang, Pan, Wang, Chen, Jin, Chen, Lu, Zhou, Chen, Ye, Wang, Yu, Zhou,
  Pan, Li, Zhou, Wu, Yun, Pei, Sun, Wang, Zeng, Liu, Liang, Gao, Yu, Zhang,
  Xiao, An, Liu, Wang, Chen, Nie, Cheng, Liu, Xie, Liu, Yang, Li, Su, Lin, Li,
  Jin, Shen, Chen, Sun, Wang, Song, Zhou, Wang, Shan, Li, Wang, Wei, Zhang, Xu,
  Li, Zhao, Sun, Wang, Yu, Zhang, Shi, Xiong, He, Piao, Wang, Tan, Ma, Liu,
  Guo, Ou, Wang, Gong, Zou, He, Xiong, Luo, You, Liu, Zhou, Zhu, Huang, Li,
  Zheng, Zhu, Ma, Tang, Zha, Yan, Ren, Ren, Sha, Fu, Xu, Xie, Zhang, Hao, Ma,
  Yan, Wu, Gu, Zhu, Liu, Li, Xie, Song, Pan, Huang, Xu, Zhang, and
  Zhang]{Guo_2025}
Daya Guo, Dejian Yang, Haowei Zhang, Junxiao Song, Peiyi Wang, Qihao Zhu,
  Runxin Xu, Ruoyu Zhang, Shirong Ma, Xiao Bi, Xiaokang Zhang, Xingkai Yu,
  Yu~Wu, Z.~F. Wu, Zhibin Gou, Zhihong Shao, Zhuoshu Li, Ziyi Gao, Aixin Liu,
  Bing Xue, Bingxuan Wang, Bochao Wu, Bei Feng, Chengda Lu, Chenggang Zhao,
  Chengqi Deng, Chong Ruan, Damai Dai, Deli Chen, Dongjie Ji, Erhang Li,
  Fangyun Lin, Fucong Dai, Fuli Luo, Guangbo Hao, Guanting Chen, Guowei Li,
  H.~Zhang, Hanwei Xu, Honghui Ding, Huazuo Gao, Hui Qu, Hui Li, Jianzhong Guo,
  Jiashi Li, Jingchang Chen, Jingyang Yuan, Jinhao Tu, Junjie Qiu, Junlong Li,
  J.~L. Cai, Jiaqi Ni, Jian Liang, Jin Chen, Kai Dong, Kai Hu, Kaichao You,
  Kaige Gao, Kang Guan, Kexin Huang, Kuai Yu, Lean Wang, Lecong Zhang, Liang
  Zhao, Litong Wang, Liyue Zhang, Lei Xu, Leyi Xia, Mingchuan Zhang, Minghua
  Zhang, Minghui Tang, Mingxu Zhou, Meng Li, Miaojun Wang, Mingming Li, Ning
  Tian, Panpan Huang, Peng Zhang, Qiancheng Wang, Qinyu Chen, Qiushi Du, Ruiqi
  Ge, Ruisong Zhang, Ruizhe Pan, Runji Wang, R.~J. Chen, R.~L. Jin, Ruyi Chen,
  Shanghao Lu, Shangyan Zhou, Shanhuang Chen, Shengfeng Ye, Shiyu Wang,
  Shuiping Yu, Shunfeng Zhou, Shuting Pan, S.~S. Li, Shuang Zhou, Shaoqing Wu,
  Tao Yun, Tian Pei, Tianyu Sun, T.~Wang, Wangding Zeng, Wen Liu, Wenfeng
  Liang, Wenjun Gao, Wenqin Yu, Wentao Zhang, W.~L. Xiao, Wei An, Xiaodong Liu,
  Xiaohan Wang, Xiaokang Chen, Xiaotao Nie, Xin Cheng, Xin Liu, Xin Xie,
  Xingchao Liu, Xinyu Yang, Xinyuan Li, Xuecheng Su, Xuheng Lin, X.~Q. Li,
  Xiangyue Jin, Xiaojin Shen, Xiaosha Chen, Xiaowen Sun, Xiaoxiang Wang, Xinnan
  Song, Xinyi Zhou, Xianzu Wang, Xinxia Shan, Y.~K. Li, Y.~Q. Wang, Y.~X. Wei,
  Yang Zhang, Yanhong Xu, Yao Li, Yao Zhao, Yaofeng Sun, Yaohui Wang, Yi~Yu,
  Yichao Zhang, Yifan Shi, Yiliang Xiong, Ying He, Yishi Piao, Yisong Wang,
  Yixuan Tan, Yiyang Ma, Yiyuan Liu, Yongqiang Guo, Yuan Ou, Yuduan Wang, Yue
  Gong, Yuheng Zou, Yujia He, Yunfan Xiong, Yuxiang Luo, Yuxiang You, Yuxuan
  Liu, Yuyang Zhou, Y.~X. Zhu, Yanping Huang, Yaohui Li, Yi~Zheng, Yuchen Zhu,
  Yunxian Ma, Ying Tang, Yukun Zha, Yuting Yan, Z.~Z. Ren, Zehui Ren, Zhangli
  Sha, Zhe Fu, Zhean Xu, Zhenda Xie, Zhengyan Zhang, Zhewen Hao, Zhicheng Ma,
  Zhigang Yan, Zhiyu Wu, Zihui Gu, Zijia Zhu, Zijun Liu, Zilin Li, Ziwei Xie,
  Ziyang Song, Zizheng Pan, Zhen Huang, Zhipeng Xu, Zhongyu Zhang, and Zhen
  Zhang.
\newblock Deepseek-r1 incentivizes reasoning in llms through reinforcement
  learning.
\newblock \emph{Nature}, 645\penalty0 (8081):\penalty0 633–638, September
  2025.
\newblock ISSN 1476-4687.
\newblock \doi{10.1038/s41586-025-09422-z}.
\newblock URL \url{http://dx.doi.org/10.1038/s41586-025-09422-z}.

\bibitem[Hendrycks et~al.(2021)Hendrycks, Burns, Kadavath, Arora, Basart, Tang,
  Song, and Steinhardt]{hendrycks2021MATH}
Dan Hendrycks, Collin Burns, Saurav Kadavath, Akul Arora, Steven Basart, Eric
  Tang, Dawn Song, and Jacob Steinhardt.
\newblock Measuring mathematical problem solving with the math dataset, 2021.
\newblock URL \url{https://arxiv.org/abs/2103.03874}.

\bibitem[Huang et~al.(2025)Huang, Liu, Zhang, Yu, and
  Li]{huang2025offlinetoonlinereinforcementlearningclassifierfree}
Xiao Huang, Xu~Liu, Enze Zhang, Tong Yu, and Shuai Li.
\newblock Offline-to-online reinforcement learning with classifier-free
  diffusion generation, 2025.
\newblock URL \url{https://arxiv.org/abs/2508.06806}.

\bibitem[Jiang and Li(2016)]{jiang2016doublyrobustoffpolicyvalue}
Nan Jiang and Lihong Li.
\newblock Doubly robust off-policy value evaluation for reinforcement learning,
  2016.
\newblock URL \url{https://arxiv.org/abs/1511.03722}.

\bibitem[Jin et~al.(2025{\natexlab{a}})Jin, Luan, Lyu, Rabusseau, Rabbany,
  Precup, and Hamdaqa]{jin2025rlfinetuninghealsood}
Hangzhan Jin, Sitao Luan, Sicheng Lyu, Guillaume Rabusseau, Reihaneh Rabbany,
  Doina Precup, and Mohammad Hamdaqa.
\newblock Rl fine-tuning heals ood forgetting in sft, 2025{\natexlab{a}}.
\newblock URL \url{https://arxiv.org/abs/2509.12235}.

\bibitem[Jin et~al.(2025{\natexlab{b}})Jin, Lv, Wu, and
  Hamdaqa]{jin2025rlpanacea}
Hangzhan Jin, Sicheng Lv, Sifan Wu, and Mohammad Hamdaqa.
\newblock Rl is neither a panacea nor a mirage: Understanding supervised vs.
  reinforcement learning fine-tuning for llms, 2025{\natexlab{b}}.
\newblock URL \url{https://arxiv.org/abs/2508.16546}.

\bibitem[Kang et~al.(2025)Kang, Kuchnik, Padthe, Vlastelica, Jia, Wu, and
  Ardalani]{kang2025quagmiressftrlposttraininghigh}
Feiyang Kang, Michael Kuchnik, Karthik Padthe, Marin Vlastelica, Ruoxi Jia,
  Carole-Jean Wu, and Newsha Ardalani.
\newblock Quagmires in sft-rl post-training: When high sft scores mislead and
  what to use instead, 2025.
\newblock URL \url{https://arxiv.org/abs/2510.01624}.

\bibitem[Lanchantin et~al.(2025)Lanchantin, Chen, Lan, Li, Saha, Wang, Xu, Yu,
  Yuan, Weston, Sukhbaatar, and
  Kulikov]{lanchantin2025bridgingofflineonlinereinforcement}
Jack Lanchantin, Angelica Chen, Janice Lan, Xian Li, Swarnadeep Saha, Tianlu
  Wang, Jing Xu, Ping Yu, Weizhe Yuan, Jason~E Weston, Sainbayar Sukhbaatar,
  and Ilia Kulikov.
\newblock Bridging offline and online reinforcement learning for llms, 2025.
\newblock URL \url{https://arxiv.org/abs/2506.21495}.

\bibitem[Lee et~al.(2021)Lee, Seo, Lee, Abbeel, and
  Shin]{lee2021offlinetoonlinereinforcementlearningbalanced}
Seunghyun Lee, Younggyo Seo, Kimin Lee, Pieter Abbeel, and Jinwoo Shin.
\newblock Offline-to-online reinforcement learning via balanced replay and
  pessimistic q-ensemble, 2021.
\newblock URL \url{https://arxiv.org/abs/2107.00591}.

\bibitem[Levine et~al.(2020)Levine, Kumar, Tucker, and
  Fu]{levine2020offlinereinforcementlearningtutorial}
Sergey Levine, Aviral Kumar, George Tucker, and Justin Fu.
\newblock Offline reinforcement learning: Tutorial, review, and perspectives on
  open problems, 2020.
\newblock URL \url{https://arxiv.org/abs/2005.01643}.

\bibitem[Lewkowycz et~al.(2022)Lewkowycz, Andreassen, Dohan, Dyer, Michalewski,
  Ramasesh, Slone, Anil, Schlag, Gutman-Solo, Wu, Neyshabur, Gur-Ari, and
  Misra]{lewkowycz2022MINERVA}
Aitor Lewkowycz, Anders Andreassen, David Dohan, Ethan Dyer, Henryk
  Michalewski, Vinay Ramasesh, Ambrose Slone, Cem Anil, Imanol Schlag, Theo
  Gutman-Solo, Yuhuai Wu, Behnam Neyshabur, Guy Gur-Ari, and Vedant Misra.
\newblock Solving quantitative reasoning problems with language models, 2022.
\newblock URL \url{https://arxiv.org/abs/2206.14858}.

\bibitem[Li et~al.(2025{\natexlab{a}})Li, Qiu, Chen, Ji, and
  Tong]{li2025beyond}
Gaotang Li, Ruizhong Qiu, Xiusi Chen, Heng Ji, and Hanghang Tong.
\newblock Beyond log likelihood: Probability-based objectives for supervised
  fine-tuning across the model capability continuum, 2025{\natexlab{a}}.
\newblock URL \url{https://arxiv.org/abs/2510.00526}.

\bibitem[Li et~al.(2025{\natexlab{b}})Li, Zhou, Lam, Yang, and
  Lu]{li2025reporeplayenhancedpolicyoptimization}
Siheng Li, Zhanhui Zhou, Wai Lam, Chao Yang, and Chaochao Lu.
\newblock Repo: Replay-enhanced policy optimization, 2025{\natexlab{b}}.
\newblock URL \url{https://arxiv.org/abs/2506.09340}.

\bibitem[Lin et~al.(2025)Lin, Wang, Qian, Wang, Srinivasan, Zeng, Jiao, Zhou,
  Gesi, Wang, Guo, Zhong, Zhang, Sanghavi, Chen, Yun, and
  Li]{lin2025sftdoesnthurtgeneral}
Jiacheng Lin, Zhongruo Wang, Kun Qian, Tian Wang, Arvind Srinivasan, Hansi
  Zeng, Ruochen Jiao, Xie Zhou, Jiri Gesi, Dakuo Wang, Yufan Guo, Kai Zhong,
  Weiqi Zhang, Sujay Sanghavi, Changyou Chen, Hyokun Yun, and Lihong Li.
\newblock Sft doesn't always hurt general capabilities: Revisiting
  domain-specific fine-tuning in llms, 2025.
\newblock URL \url{https://arxiv.org/abs/2509.20758}.

\bibitem[Liu et~al.(2025)Liu, Fan, Jiang, Ding, Hu, Zhang, Shi, Weng, Chen,
  Chen, Huang, Zhang, Zhao, Yan, and He]{liu2025synlogic}
Junteng Liu, Yuanxiang Fan, Zhuo Jiang, Han Ding, Yongyi Hu, Chi Zhang, Yiqi
  Shi, Shitong Weng, Aili Chen, Shiqi Chen, Yunan Huang, Mozhi Zhang, Pengyu
  Zhao, Junjie Yan, and Junxian He.
\newblock Synlogic: Synthesizing verifiable reasoning data at scale for
  learning logical reasoning and beyond, 2025.
\newblock URL \url{https://arxiv.org/abs/2505.19641}.

\bibitem[Liu et~al.(2018)Liu, Li, Tang, and
  Zhou]{liu2018breakingcursehorizoninfinitehorizon}
Qiang Liu, Lihong Li, Ziyang Tang, and Dengyong Zhou.
\newblock Breaking the curse of horizon: Infinite-horizon off-policy
  estimation, 2018.
\newblock URL \url{https://arxiv.org/abs/1810.12429}.

\bibitem[Liu et~al.(2019)Liu, Swaminathan, Agarwal, and
  Brunskill]{liu2019offpolicypolicygradientstate}
Yao Liu, Adith Swaminathan, Alekh Agarwal, and Emma Brunskill.
\newblock Off-policy policy gradient with state distribution correction, 2019.
\newblock URL \url{https://arxiv.org/abs/1904.08473}.

\bibitem[Liu et~al.(2020)Liu, Bacon, and
  Brunskill]{liu2020understandingcursehorizonoffpolicy}
Yao Liu, Pierre-Luc Bacon, and Emma Brunskill.
\newblock Understanding the curse of horizon in off-policy evaluation via
  conditional importance sampling, 2020.
\newblock URL \url{https://arxiv.org/abs/1910.06508}.

\bibitem[Mehta et~al.(2024)Mehta, Ciftci, Ramachandran, Bansal, and
  Losey]{mehta2024stablebccontrollingcovariateshift}
Shaunak~A. Mehta, Yusuf~Umut Ciftci, Balamurugan Ramachandran, Somil Bansal,
  and Dylan~P. Losey.
\newblock Stable-bc: Controlling covariate shift with stable behavior cloning,
  2024.
\newblock URL \url{https://arxiv.org/abs/2408.06246}.

\bibitem[Metelli et~al.(2020)Metelli, Papini, Montali, and
  Restelli]{metelli2020ispo}
Alberto~Maria Metelli, Matteo Papini, Nico Montali, and Marcello Restelli.
\newblock Importance sampling techniques for policy optimization.
\newblock \emph{Journal of Machine Learning Research}, 21\penalty0
  (141):\penalty0 1--75, 2020.
\newblock URL \url{http://jmlr.org/papers/v21/20-124.html}.

\bibitem[Mukherjee et~al.(2025{\natexlab{a}})Mukherjee, Yuan, Hakkani-Tur, and
  Peng]{mukherjee2025reinforcementlearningfinetunessmall}
Sagnik Mukherjee, Lifan Yuan, Dilek Hakkani-Tur, and Hao Peng.
\newblock Reinforcement learning finetunes small subnetworks in large language
  models, 2025{\natexlab{a}}.
\newblock URL \url{https://arxiv.org/abs/2505.11711}.

\bibitem[Mukherjee et~al.(2025{\natexlab{b}})Mukherjee, Lai, Addanki, Rossi,
  Yoon, Bui, Rao, Subramanian, and
  Kveton]{mukherjee2025offlinerlrewardweightedfinetuning}
Subhojyoti Mukherjee, Viet~Dac Lai, Raghavendra Addanki, Ryan Rossi, Seunghyun
  Yoon, Trung Bui, Anup Rao, Jayakumar Subramanian, and Branislav Kveton.
\newblock Offline rl by reward-weighted fine-tuning for conversation
  optimization, 2025{\natexlab{b}}.
\newblock URL \url{https://arxiv.org/abs/2506.06964}.

\bibitem[Nachum et~al.(2019)Nachum, Chow, Dai, and
  Li]{nachum2019dualdicebehavioragnosticestimationdiscounted}
Ofir Nachum, Yinlam Chow, Bo~Dai, and Lihong Li.
\newblock Dualdice: Behavior-agnostic estimation of discounted stationary
  distribution corrections, 2019.
\newblock URL \url{https://arxiv.org/abs/1906.04733}.

\bibitem[Precup et~al.(2000)Precup, Sutton, and Singh]{pdis2000}
Doina Precup, Richard~S. Sutton, and Satinder~P. Singh.
\newblock Eligibility traces for off-policy policy evaluation.
\newblock In \emph{Proceedings of the Seventeenth International Conference on
  Machine Learning}, ICML '00, page 759–766, San Francisco, CA, USA, 2000.
  Morgan Kaufmann Publishers Inc.
\newblock ISBN 1558607072.

\bibitem[{Prime Intellect}(2025)]{primeintellect_synthetic2_2025}
{Prime Intellect}.
\newblock {SYNTHETIC-2}, 2025.
\newblock URL \url{https://huggingface.co/datasets/PrimeIntellect/SYNTHETIC-2}.
\newblock Updated Oct 7, 2025. Accessed Jan 18, 2026.

\bibitem[Qian et~al.(2025)Qian, Acikgoz, He, Wang, Chen, Hakkani-Tür, Tur, and
  Ji]{qian2025toolrlrewardtoollearning}
Cheng Qian, Emre~Can Acikgoz, Qi~He, Hongru Wang, Xiusi Chen, Dilek
  Hakkani-Tür, Gokhan Tur, and Heng Ji.
\newblock Toolrl: Reward is all tool learning needs, 2025.
\newblock URL \url{https://arxiv.org/abs/2504.13958}.

\bibitem[Qin and Springenberg(2025)]{qin2025iwsft}
Chongli Qin and Jost~Tobias Springenberg.
\newblock Supervised fine tuning on curated data is reinforcement learning (and
  can be improved), 2025.
\newblock URL \url{https://arxiv.org/abs/2507.12856}.

\bibitem[Richemond et~al.(2024)Richemond, Tang, Guo, Calandriello, Azar,
  Rafailov, Pires, Tarassov, Spangher, Ellsworth, Severyn, Mallinson, Shani,
  Shamir, Joshi, Liu, Munos, and
  Piot]{richemond2024offlineregularisedreinforcementlearning}
Pierre~Harvey Richemond, Yunhao Tang, Daniel Guo, Daniele Calandriello,
  Mohammad~Gheshlaghi Azar, Rafael Rafailov, Bernardo~Avila Pires, Eugene
  Tarassov, Lucas Spangher, Will Ellsworth, Aliaksei Severyn, Jonathan
  Mallinson, Lior Shani, Gil Shamir, Rishabh Joshi, Tianqi Liu, Remi Munos, and
  Bilal Piot.
\newblock Offline regularised reinforcement learning for large language models
  alignment, 2024.
\newblock URL \url{https://arxiv.org/abs/2405.19107}.

\bibitem[Ross and
  Bagnell(2014)]{ross2014reinforcementimitationlearninginteractive}
Stephane Ross and J.~Andrew Bagnell.
\newblock Reinforcement and imitation learning via interactive no-regret
  learning, 2014.
\newblock URL \url{https://arxiv.org/abs/1406.5979}.

\bibitem[Ross et~al.(2011)Ross, Gordon, and Bagnell]{ross2011imitation}
Stephane Ross, Geoffrey~J. Gordon, and J.~Andrew Bagnell.
\newblock A reduction of imitation learning and structured prediction to
  no-regret online learning, 2011.
\newblock URL \url{https://arxiv.org/abs/1011.0686}.

\bibitem[Rowland et~al.(2020)Rowland, Harutyunyan, van Hasselt, Borsa, Schaul,
  Munos, and Dabney]{cis_off_policy}
Mark Rowland, Anna Harutyunyan, Hado van Hasselt, Diana Borsa, Tom Schaul, Remi
  Munos, and Will Dabney.
\newblock Conditional importance sampling for off-policy learning.
\newblock In Silvia Chiappa and Roberto Calandra, editors, \emph{Proceedings of
  the Twenty Third International Conference on Artificial Intelligence and
  Statistics}, volume 108 of \emph{Proceedings of Machine Learning Research},
  pages 45--55. PMLR, 26--28 Aug 2020.
\newblock URL \url{https://proceedings.mlr.press/v108/rowland20b.html}.

\bibitem[Sanyal et~al.(2025)Sanyal, Prairie, Das, Kavis, and
  Sanghavi]{sanyal2025upweightingeasysamplesfinetuning}
Sunny Sanyal, Hayden Prairie, Rudrajit Das, Ali Kavis, and Sujay Sanghavi.
\newblock Upweighting easy samples in fine-tuning mitigates forgetting, 2025.
\newblock URL \url{https://arxiv.org/abs/2502.02797}.

\bibitem[Shao et~al.(2024)Shao, Wang, Zhu, Xu, Song, Bi, Zhang, Zhang, Li, Wu,
  and Guo]{shao2024deepseekmath}
Zhihong Shao, Peiyi Wang, Qihao Zhu, Runxin Xu, Junxiao Song, Xiao Bi, Haowei
  Zhang, Mingchuan Zhang, Y.~K. Li, Y.~Wu, and Daya Guo.
\newblock Deepseekmath: Pushing the limits of mathematical reasoning in open
  language models, 2024.
\newblock URL \url{https://arxiv.org/abs/2402.03300}.

\bibitem[Shenfeld et~al.(2025)Shenfeld, Pari, and Agrawal]{shenfeld2025rlrazor}
Idan Shenfeld, Jyothish Pari, and Pulkit Agrawal.
\newblock Rl's razor: Why online reinforcement learning forgets less, 2025.
\newblock URL \url{https://arxiv.org/abs/2509.04259}.

\bibitem[Snell et~al.(2023)Snell, Kostrikov, Su, Yang, and
  Levine]{snell2023offlinerlnaturallanguage}
Charlie Snell, Ilya Kostrikov, Yi~Su, Mengjiao Yang, and Sergey Levine.
\newblock Offline rl for natural language generation with implicit language q
  learning, 2023.
\newblock URL \url{https://arxiv.org/abs/2206.11871}.

\bibitem[Sun et~al.(2026)Sun, Cai, He, Chen, Bao, Yang, Wu, and
  Wang]{sun2026distributionalclarityhiddendriver}
Shaoning Sun, Mingzhu Cai, Huang He, Bingjin Chen, Siqi Bao, Yujiu Yang, Hua
  Wu, and Haifeng Wang.
\newblock Distributional clarity: The hidden driver of rl-friendliness in large
  language models, 2026.
\newblock URL \url{https://arxiv.org/abs/2601.06911}.

\bibitem[Sun et~al.(2017)Sun, Venkatraman, Gordon, Boots, and
  Bagnell]{sun2017deeplyaggrevateddifferentiableimitation}
Wen Sun, Arun Venkatraman, Geoffrey~J. Gordon, Byron Boots, and J.~Andrew
  Bagnell.
\newblock Deeply aggrevated: Differentiable imitation learning for sequential
  prediction, 2017.
\newblock URL \url{https://arxiv.org/abs/1703.01030}.

\bibitem[Sutton and Barto(2018)]{sutoonrl}
Richard~S. Sutton and Andrew~G. Barto.
\newblock \emph{Reinforcement Learning: An Introduction}.
\newblock A Bradford Book, Cambridge, MA, USA, 2018.
\newblock ISBN 0262039249.

\bibitem[Thomas and
  Brunskill(2016)]{thomas2016dataefficientoffpolicypolicyevaluation}
Philip~S. Thomas and Emma Brunskill.
\newblock Data-efficient off-policy policy evaluation for reinforcement
  learning, 2016.
\newblock URL \url{https://arxiv.org/abs/1604.00923}.

\bibitem[Uehara et~al.(2022)Uehara, Shi, and Kallus]{uehara2022opesurvey}
Masatoshi Uehara, Chengchun Shi, and Nathan Kallus.
\newblock A review of off-policy evaluation in reinforcement learning, 2022.
\newblock URL \url{https://arxiv.org/abs/2212.06355}.

\bibitem[Wang et~al.(2024)Wang, Hao, Dong, Zhang, Bao, Yang, and
  Wu]{wang2024oreo}
Huaijie Wang, Shibo Hao, Hanze Dong, Shenao Zhang, Yilin Bao, Ziran Yang, and
  Yi~Wu.
\newblock Offline reinforcement learning for llm multi-step reasoning, 2024.
\newblock URL \url{https://arxiv.org/abs/2412.16145}.

\bibitem[Wei et~al.(2025)Wei, Duchenne, Copet, Carbonneaux, Zhang, Fried,
  Synnaeve, Singh, and Wang]{wei2025swerladvancingllmreasoning}
Yuxiang Wei, Olivier Duchenne, Jade Copet, Quentin Carbonneaux, Lingming Zhang,
  Daniel Fried, Gabriel Synnaeve, Rishabh Singh, and Sida~I. Wang.
\newblock Swe-rl: Advancing llm reasoning via reinforcement learning on open
  software evolution, 2025.
\newblock URL \url{https://arxiv.org/abs/2502.18449}.

\bibitem[Wu et~al.(2025)Wu, Zhou, Ziheng, Peng, Ye, Hu, Zhu, Qi, Yang, and
  Yang]{wu2025dft}
Yongliang Wu, Yizhou Zhou, Zhou Ziheng, Yingzhe Peng, Xinyu Ye, Xinting Hu,
  Wenbo Zhu, Lu~Qi, Ming-Hsuan Yang, and Xu~Yang.
\newblock On the generalization of sft: A reinforcement learning perspective
  with reward rectification, 2025.
\newblock URL \url{https://arxiv.org/abs/2508.05629}.

\bibitem[Yang et~al.(2024)Yang, Zhang, Hui, Gao, Yu, Li, Liu, Tu, Zhou, Lin,
  Lu, Xue, Lin, Liu, Ren, and Zhang]{yang2024qwen25}
An~Yang, Beichen Zhang, Binyuan Hui, Bofei Gao, Bowen Yu, Chengpeng Li,
  Dayiheng Liu, Jianhong Tu, Jingren Zhou, Junyang Lin, Keming Lu, Mingfeng
  Xue, Runji Lin, Tianyu Liu, Xingzhang Ren, and Zhenru Zhang.
\newblock Qwen2.5-math technical report: Toward mathematical expert model via
  self-improvement, 2024.
\newblock URL \url{https://arxiv.org/abs/2409.12122}.

\bibitem[Yang et~al.(2025)Yang, Li, Yang, Zhang, Hui, Zheng, Yu, Gao, Huang,
  Lv, Zheng, Liu, Zhou, Huang, Hu, Ge, Wei, Lin, Tang, Yang, Tu, Zhang, Yang,
  Yang, Zhou, Zhou, Lin, Dang, Bao, Yang, Yu, Deng, Li, Xue, Li, Zhang, Wang,
  Zhu, Men, Gao, Liu, Luo, Li, Tang, Yin, Ren, Wang, Zhang, Ren, Fan, Su,
  Zhang, Zhang, Wan, Liu, Wang, Cui, Zhang, Zhou, and
  Qiu]{yang2025qwen3technicalreport}
An~Yang, Anfeng Li, Baosong Yang, Beichen Zhang, Binyuan Hui, Bo~Zheng, Bowen
  Yu, Chang Gao, Chengen Huang, Chenxu Lv, Chujie Zheng, Dayiheng Liu, Fan
  Zhou, Fei Huang, Feng Hu, Hao Ge, Haoran Wei, Huan Lin, Jialong Tang, Jian
  Yang, Jianhong Tu, Jianwei Zhang, Jianxin Yang, Jiaxi Yang, Jing Zhou,
  Jingren Zhou, Junyang Lin, Kai Dang, Keqin Bao, Kexin Yang, Le~Yu, Lianghao
  Deng, Mei Li, Mingfeng Xue, Mingze Li, Pei Zhang, Peng Wang, Qin Zhu, Rui
  Men, Ruize Gao, Shixuan Liu, Shuang Luo, Tianhao Li, Tianyi Tang, Wenbiao
  Yin, Xingzhang Ren, Xinyu Wang, Xinyu Zhang, Xuancheng Ren, Yang Fan, Yang
  Su, Yichang Zhang, Yinger Zhang, Yu~Wan, Yuqiong Liu, Zekun Wang, Zeyu Cui,
  Zhenru Zhang, Zhipeng Zhou, and Zihan Qiu.
\newblock Qwen3 technical report, 2025.
\newblock URL \url{https://arxiv.org/abs/2505.09388}.

\bibitem[Yu et~al.(2025)Yu, Zhang, Zhu, Yuan, Zuo, Yue, Dai, Fan, Liu, Liu,
  Liu, Lin, Lin, Ma, Sheng, Tong, Zhang, Zhang, Zhang, Zhu, Zhu, Chen, Chen,
  Wang, Yu, Song, Wei, Zhou, Liu, Ma, Zhang, Yan, Qiao, Wu, and
  Wang]{yu2025dapo}
Qiying Yu, Zheng Zhang, Ruofei Zhu, Yufeng Yuan, Xiaochen Zuo, Yu~Yue, Weinan
  Dai, Tiantian Fan, Gaohong Liu, Lingjun Liu, Xin Liu, Haibin Lin, Zhiqi Lin,
  Bole Ma, Guangming Sheng, Yuxuan Tong, Chi Zhang, Mofan Zhang, Wang Zhang,
  Hang Zhu, Jinhua Zhu, Jiaze Chen, Jiangjie Chen, Chengyi Wang, Hongli Yu,
  Yuxuan Song, Xiangpeng Wei, Hao Zhou, Jingjing Liu, Wei-Ying Ma, Ya-Qin
  Zhang, Lin Yan, Mu~Qiao, Yonghui Wu, and Mingxuan Wang.
\newblock Dapo: An open-source llm reinforcement learning system at scale,
  2025.
\newblock URL \url{https://arxiv.org/abs/2503.14476}.

\bibitem[Yue et~al.(2025)Yue, Chen, Lu, Zhao, Wang, Yue, Song, and
  Huang]{yue2025doesreinforcementlearningreally}
Yang Yue, Zhiqi Chen, Rui Lu, Andrew Zhao, Zhaokai Wang, Yang Yue, Shiji Song,
  and Gao Huang.
\newblock Does reinforcement learning really incentivize reasoning capacity in
  llms beyond the base model?, 2025.
\newblock URL \url{https://arxiv.org/abs/2504.13837}.

\bibitem[Zhang et~al.(2025{\natexlab{a}})Zhang, Xie, Sun, Chen, Wang, Li, Ding,
  and Zhou]{zhang2025onpolicyrlmeetsoffpolicy}
Wenhao Zhang, Yuexiang Xie, Yuchang Sun, Yanxi Chen, Guoyin Wang, Yaliang Li,
  Bolin Ding, and Jingren Zhou.
\newblock On-policy rl meets off-policy experts: Harmonizing supervised
  fine-tuning and reinforcement learning via dynamic weighting,
  2025{\natexlab{a}}.
\newblock URL \url{https://arxiv.org/abs/2508.11408}.

\bibitem[Zhang et~al.(2025{\natexlab{b}})Zhang, Feng, Guan, He, and
  Wu]{zhang2025beyondonline}
Zhang Zhang, Guhao Feng, Jian Guan, Di~He, and Wei Wu.
\newblock Beyond online sampling: Bridging offline-to-online alignment via
  dynamic data transformation for {LLM}s.
\newblock In Christos Christodoulopoulos, Tanmoy Chakraborty, Carolyn Rose, and
  Violet Peng, editors, \emph{Proceedings of the 2025 Conference on Empirical
  Methods in Natural Language Processing}, pages 27097--27109, Suzhou, China,
  November 2025{\natexlab{b}}. Association for Computational Linguistics.
\newblock ISBN 979-8-89176-332-6.
\newblock \doi{10.18653/v1/2025.emnlp-main.1378}.
\newblock URL \url{https://aclanthology.org/2025.emnlp-main.1378/}.

\bibitem[Zhao et~al.(2025)Zhao, Meterez, Kakade, Pehlevan, Jelassi, and
  Malach]{zhao2025echochamber}
Rosie Zhao, Alexandru Meterez, Sham Kakade, Cengiz Pehlevan, Samy Jelassi, and
  Eran Malach.
\newblock Echo chamber: Rl post-training amplifies behaviors learned in
  pretraining, 2025.
\newblock URL \url{https://arxiv.org/abs/2504.07912}.

\bibitem[Zhao et~al.(2022)Zhao, Boney, Ilin, Kannala, and
  Pajarinen]{zhao2022adaptivebehaviorcloningregularization}
Yi~Zhao, Rinu Boney, Alexander Ilin, Juho Kannala, and Joni Pajarinen.
\newblock Adaptive behavior cloning regularization for stable offline-to-online
  reinforcement learning, 2022.
\newblock URL \url{https://arxiv.org/abs/2210.13846}.

\bibitem[Zhu et~al.(2025{\natexlab{a}})Zhu, Zhang, Huang, Su, Liu, Zhao,
  Fedorov, Pirsiavash, Sha, Lee, Pan, Wang, Tian, and Tai]{zhu2025pathnottaken}
Hanqing Zhu, Zhenyu Zhang, Hanxian Huang, DiJia Su, Zechun Liu, Jiawei Zhao,
  Igor Fedorov, Hamed Pirsiavash, Zhizhou Sha, Jinwon Lee, David~Z. Pan,
  Zhangyang Wang, Yuandong Tian, and Kai~Sheng Tai.
\newblock The path not taken: Rlvr provably learns off the principals,
  2025{\natexlab{a}}.
\newblock URL \url{https://arxiv.org/abs/2511.08567}.

\bibitem[Zhu et~al.(2025{\natexlab{b}})Zhu, Su, Lai, Ma, Zhang, Yang, and
  Chen]{zhu2025aft}
He~Zhu, Junyou Su, Peng Lai, Ren Ma, Wenjia Zhang, Linyi Yang, and Guanhua
  Chen.
\newblock Anchored supervised fine-tuning, 2025{\natexlab{b}}.
\newblock URL \url{https://arxiv.org/abs/2509.23753}.

\bibitem[Zhu et~al.(2025{\natexlab{c}})Zhu, Xie, Wang, Sun, Wang, and
  Liu]{zhu2025proximalsft}
Wenhong Zhu, Ruobing Xie, Rui Wang, Xingwu Sun, Di~Wang, and Pengfei Liu.
\newblock Proximal supervised fine-tuning, 2025{\natexlab{c}}.
\newblock URL \url{https://arxiv.org/abs/2508.17784}.

\bibitem[Zu et~al.(2025)Zu, Zhou, and
  Zhang]{zu2025behavioradaptiveqlearningunifyingframework}
Lipeng Zu, Hansong Zhou, and Xiaonan Zhang.
\newblock Behavior-adaptive q-learning: A unifying framework for
  offline-to-online rl, 2025.
\newblock URL \url{https://arxiv.org/abs/2511.03695}.

\end{thebibliography}

\newpage

\normalsize

\appendix

\section*{Contents}
\renewcommand{\arraystretch}{1.15}
\begin{tabular}{@{}p{0.04\linewidth}p{0.78\linewidth}@{\hspace{0.5em}}r@{}}
\textbf{A}  & Black-Box \name: Proxy and Ensemble Behavior Policies \dotfill & \pageref{app:blackbox} \\
\textbf{B}  & Hyperparameter Sensitivity: Clipping Range \dotfill & \pageref{app:hparam} \\
\textbf{C}  & Block-Size Ablation \dotfill & \pageref{app:block_size} \\
\textbf{D}  & Beyond Reasoning: Instruction-Following \dotfill & \pageref{app:instruction_following} \\
\textbf{E}  & Comparison with CHORD \dotfill & \pageref{app:chord} \\
\textbf{F}  & Does \name Suppress Useful Low-Probability Trajectories? \dotfill & \pageref{app:reasoning_patterns} \\
\textbf{G}  & Compatibility with Other RL Algorithms (DAPO) \dotfill & \pageref{app:pear_dapo} \\
\textbf{H}  & Computation of Metrics \dotfill & \pageref{app:metrics} \\
\textbf{I}  & Suffix Change-of-Measure \dotfill & \pageref{app:suffix_com} \\
\textbf{J}  & An Alternative Intuition: Suffix Ratios as Off-Policy Value Estimates \dotfill & \pageref{app:alt_intuition} \\
\textbf{K}  & Details on Baselines (TALR, Beyond Log-Likelihood) \dotfill & \pageref{app:baselines} \\
\textbf{L}  & Discussion: Should Offline Training Match Online RL Characteristics? \dotfill & \pageref{app:discussion_twostage} \\
\textbf{M}  & Offline vs Online Metrics: Detailed Tables \dotfill & \pageref{app:offline_online_tables} \\
\end{tabular}
\clearpage

\section{Black-Box \name: Proxy and Ensemble Behavior Policies}
\label{app:blackbox}

\name as presented in the main text assumes white-box access to the behavior policy $\pi_\beta$ so that token log-probabilities can be evaluated exactly. This holds whenever the offline data is generated by a known open-weight teacher (as in our main experiments) but can be too strong an assumption when SFT data is curated, scraped, or produced by a closed-source teacher. We show here that \name remains effective when $\pi_\beta$ is only available through an \emph{approximate} proxy.

\paragraph{Practical strategies.} We consider two ways to estimate $\pi_\beta$ without exact access:
(i) a \emph{single proxy} model from the same model family but with a different size or post-training (e.g.\ a different Qwen3 variant), and
(ii) an \emph{ensemble} of proxies, where we average per-token probabilities across multiple models.

\paragraph{Setup.} We use the same SynLogic setting and SFT/RL recipe as in \S\ref{sec:experiments}. The data is generated by Qwen3-8B (as in the main paper). We approximate $\pi_\beta$ using each of Qwen3-4B-Base, Qwen3-4B-Thinking, Qwen3-32B individually, and a uniform average over the three (Mixture), \emph{excluding} the actual generator. We report Pass@1/Pass@8 after identical GRPO on Qwen3-1.7B-Base and Qwen3-4B-Base; for context we also report the white-box \name baseline.

\begin{figure}[h!]
    \centering
    \includegraphics[width=0.5\linewidth]{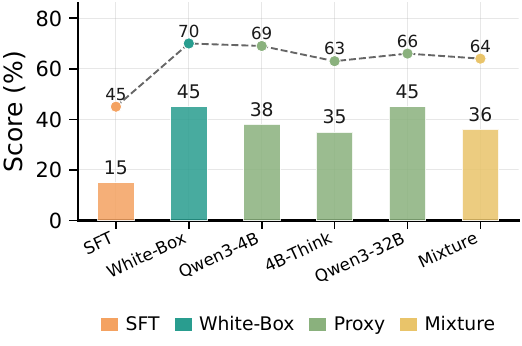}
    \caption{Black-box \name on SynLogic (Qwen3-1.7B-Base). Bars show Pass@1; dashed-line markers show Pass@8. \textsc{SFT} is the no-PEAR lower bound; \textsc{White-Box} uses the true behavior policy; remaining methods use an approximate $\pi_\beta$ (single-model proxies or the \textsc{Mixture} ensemble). Every proxy variant substantially beats SFT, and the strongest proxy (Qwen3-32B) matches white-box.}
    \label{fig:blackbox_pear}
\end{figure}

\begin{figure}[h!]
    \centering
    \includegraphics[width=0.5\linewidth]{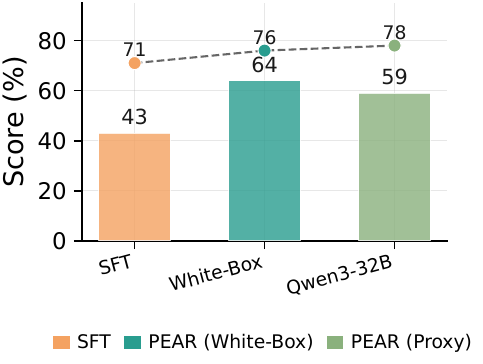}
    \caption{Black-box \name on SynLogic (Qwen3-4B-Base). PEAR with only a Qwen3-32B proxy still outperforms SFT on both metrics.}
    \label{fig:blackbox_pear_4b}
\end{figure}

\paragraph{Why does the proxy quality matter?} We compute the Pearson correlation between each proxy's per-token probabilities and those of the true behavior model across the dataset. Qwen3-4B-Thinking has the lowest correlation and correspondingly the weakest post-RL performance; Qwen3-32B has the highest correlation and best proxy performance. This supports a simple practical heuristic: choose proxies whose token-level statistics align well with the data generator.

\begin{table}[h!]
    \centering
    \small
    \begin{tabular}{lc}
        \toprule
        Proxy & Pearson w.r.t.\ behavior \\
        \midrule
        Qwen3-32B           & 0.7328 \\
        Qwen3-4B            & 0.7237 \\
        Ensemble (Mixture)  & 0.7263 \\
        Qwen3-4B-Thinking   & 0.6605 \\
        \bottomrule
    \end{tabular}
    \caption{Pearson correlation between each proxy's per-token probabilities and the true behavior model.}
    \label{tab:proxy_pearson}
\end{table}

\paragraph{Takeaway.} Across all proxy variants, \name substantially outperforms SFT+GRPO. While exact $\pi_\beta$ remains preferred when available, these results show that \name can be operationalized in the more realistic setting where the data-generating policy is only approximately accessible.

\section{Hyperparameter Sensitivity: Clipping Range}
\label{app:hparam}

\name uses log-importance clipping (\S\ref{sec:numerical_stability}) as a standard variance-reduction step. The clipping range is held fixed across all experiments in the paper; here we vary it to characterize sensitivity, on Qwen3-1.7B-Base trained on SynLogic.

We sweep $\min\in\{-0.08,-0.3\}$ and $\max\in\{0.08,0.3\}$ on the per-decision log-ratio. Pre-RL Pass@32 averages 47.5\% (std 3.95\%) and Pass@64 averages 52.19\% (std 2.58\%). Variation across clip ranges is mild and does not lead to qualitative changes in behavior.

\begin{table}[h!]
    \centering
    \small
    \begin{tabular}{lc}
        \toprule
        Clip range & Pass@32 / Pass@64 \\
        \midrule
        $[-0.08,\,0.08]$ & 42.50 / 48.75 \\
        $[-0.08,\,0.30]$ & 46.25 / 52.50 \\
        $[-0.30,\,0.08]$ & 50.00 / 52.50 \\
        $[-0.30,\,0.30]$ & 51.25 / 55.00 \\
        \bottomrule
    \end{tabular}
    \caption{Sensitivity to clipping range on Qwen3-1.7B-Base, SynLogic. Differences are within roughly one standard deviation across settings.}
    \label{tab:hparam_clip}
\end{table}

\section{Block-Size Ablation}
\label{app:block_size}

We ablate block size $B$ (\S\ref{sec:method_block_level}) on Qwen3-1.7B-Base, SynLogic, comparing the three regimes: token-level ($B=1$), block-level ($B\in\{4,8\}$), and sequence-level (single global weight).

\begin{table}[h!]
    \centering
    \small
    \begin{tabular}{lcc}
        \toprule
        Setting   & Pass@1 & Pass@8 \\
        \midrule
        Sequence  & 57.50  & 67.50 \\
        Block-4   & 60.00  & 83.75 \\
        Block-8   & 68.75  & 77.50 \\
        Block-1 (token) & 45.00 & 70.00 \\
        \bottomrule
    \end{tabular}
    \caption{Effect of granularity. Intermediate block sizes ($B\in\{4,8\}$) balance fine-grained signal against multiplicative variance and outperform both extremes.}
    \label{tab:block_size}
\end{table}

\paragraph{Discussion.} The fully fine-grained end ($B=1$) yields lower Pass@1, consistent with higher per-step variance in the multiplicative weight. The fully coarse end (Sequence) is also weakest on Pass@8: a single global weight applied to every token loses meaningful within-sequence variation. Intermediate $B$ provides a better stability--granularity trade-off, and we recommend it as a default.

\section{Beyond Reasoning: Instruction-Following}
\label{app:instruction_following}

The main experiments target verifiable reasoning. To probe whether \name's benefit extends beyond math/logic, we apply the same SFT$\rightarrow$RL pipeline to instruction-following.

\paragraph{Setup.} We use the instruction-following split of \textsc{SYNTHETIC-2}~\citep{primeintellect_synthetic2_2025} for SFT, and AI2's Dolci-RL-Zero-IF-7B dataset for RL. The student model is Qwen3-0.6B-Base. We evaluate on IF-Eval and IF-Bench, two standard instruction-following benchmarks.

\begin{table}[h!]
    \centering
    \small
    \begin{tabular}{lcc}
        \toprule
        & SFT + GRPO & \name + GRPO \\
        \midrule
        IF-Eval (Qwen3-0.6B-Base)  & 37.34\% & 57.67\% \\
        IF-Bench (Qwen3-0.6B-Base) & 16.67\% & 29.59\% \\
        \bottomrule
    \end{tabular}
    \caption{\name continues to outperform SFT+GRPO on instruction-following benchmarks by 20+ absolute points on IF-Eval, indicating the correction is not specific to math/logic.}
    \label{tab:instruction_following}
\end{table}

\paragraph{Discussion.} The gain ($+20.3$ pp on IF-Eval; $+12.9$ pp on IF-Bench) is comparable in magnitude to those observed on reasoning tasks. This is consistent with our central hypothesis: whenever the offline data generator differs from the model being optimized online, importance-weighted SFT yields a stronger initialization. Generalization to RLHF with \emph{learned} reward models is a natural next step we leave for future work.

\section{Comparison with CHORD}
\label{app:chord}

CHORD~\citep{zhang2025onpolicyrlmeetsoffpolicy} is a representative SFT/RL-hybrid method that mixes supervised and policy-gradient updates during RL. As acknowledged in Appendix~D.1 of the CHORD paper, the method is not designed to cold-start from a base model: it operates at the RL stage, after the model is already a competent policy. \name operates at the SFT stage and replaces SFT before RL. We therefore view CHORD and \name as targeting different but complementary stages of the pipeline rather than competing methods. For empirical context, we run CHORD on the same SynLogic setting using Qwen3-0.6B-Base under matched SFT and RL data.

\begin{table}[h!]
    \centering
    \small
    \begin{tabular}{lcc}
        \toprule
        Method & Pass@1 & Pass@8 \\
        \midrule
        SFT     & 10.00\% & 40.00\% \\
        CHORD   & 11.40\% & 31.25\% \\
        \name   & 12.50\% & 42.50\% \\
        \bottomrule
    \end{tabular}
    \caption{\name vs.\ CHORD on SynLogic with Qwen3-0.6B-Base. The two methods address different points of the SFT$\rightarrow$RL pipeline; results are reported for completeness.}
    \label{tab:chord}
\end{table}

\section{Does \name Suppress Useful Low-Probability Trajectories?}
\label{app:reasoning_patterns}

A natural concern with any reweighting scheme is whether it inadvertently down-weights tokens whose ``low'' weight reflects genuine novelty rather than implausibility (e.g.\ formats, exploratory tokens). \name weights by \emph{future plausibility} (suffix continuation), not by the current token's own likelihood, so the construction itself is self-correcting. We add three pieces of empirical evidence.

\paragraph{(1) \name preserves reasoning patterns.} We count occurrences of common reasoning-pattern words (``wait'', ``hmm'', ``alternatively'', ``okay'', ``therefore'') in generations from SFT- and \name-trained models on the same evaluation set. The counts are comparable; \name does not collapse reasoning patterns.

\begin{table}[h!]
    \centering
    \small
    \begin{tabular}{lccccc}
        \toprule
        Method & wait & hmm & alternatively & okay & therefore \\
        \midrule
        SFT   & 1691 & 76 & 234 & 176 & 300 \\
        \name & 1540 & 74 & 267 & 158 & 244 \\
        \bottomrule
    \end{tabular}
    \caption{Reasoning-pattern token counts. \name maintains comparable usage of standard reasoning markers.}
    \label{tab:reasoning_patterns}
\end{table}

\paragraph{(2) \name does not hurt learning of required answer formats.} Standard answer formats (e.g.\ \texttt{\textbackslash boxed\{\}}) are reliably produced by \name-initialized models with no observable degradation in format compliance.

\paragraph{(3) Cross-domain transfer.} \name's gains transfer across task distributions: e.g.\ offline training on SynLogic followed by RL on Enigmata still outperforms SFT initialization (Table~\ref{tab:synlogic_enigmata} in the main paper). If \name suppressed useful structure, we would expect such transfer to break.

Together, these results indicate that \name's reweighting preserves rather than discards informative low-probability content.

\section{Compatibility with Other RL Algorithms (DAPO)}
\label{app:pear_dapo}
\begin{figure}[h!]
    \centering
    \includegraphics[width=0.5\linewidth]{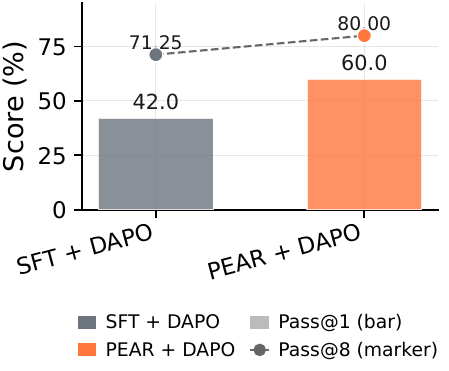}
    \caption{\name continues to outperform SFT when GRPO is replaced with DAPO~\citep{yu2025dapo}, another widely used RLVR algorithm. Same Qwen3-1.7B-Base/SynLogic setup as \S\ref{sec:experiments}.}
    \label{fig:pear_dapo}
\end{figure}
To check that \name's benefit is not GRPO-specific, we replace GRPO with DAPO~\citep{yu2025dapo} and rerun the same Qwen3-1.7B-Base/SynLogic recipe used throughout \S\ref{sec:experiments}. Figure~\ref{fig:pear_dapo} shows \name+DAPO outperforms SFT+DAPO by $+18$ Pass@1 and $+8.75$ Pass@8, mirroring the GRPO gains. The offline-to-online mismatch correction transfers across RLVR algorithms rather than being tied to GRPO.

\section{Computation Of Metrics}
\label{app:metrics}
RLVR for LLMs has become increasingly expensive since the model needs to rollout on the training set and get updates at the same time. This gets even worse when the RLVR environment involves tool calling or code execution \citep{wei2025swerladvancingllmreasoning, qian2025toolrlrewardtoollearning}, which can take minutes to hours to finish. Therefore, understanding the signals that can suggest a model's potential after RLVR can save a huge amount of compute, and has received great attention in the LLM community \citep{sun2026distributionalclarityhiddendriver}. In this section, we discuss the implementation details of the signal metrics we evaluated.

\subsection{KL}
We compute the forward KL between the base model and the model trained on offline data to understand how much the trained model's distribution diverge from the original one, and whether big or small divergence result in superior performance. Specifically, the metric we evaluated is defined as follows:
\begin{align*}
    \text{Forward KL}=\text{KL}(P_{\text{base}}||P_{\text{trained}})=\Sigma P_{\text{base}}\log{(P_\text{base}/P_\text{trained})}
\end{align*}

To evaluate the forward KL, we first collect a calibration set of 269 question-answer pairs in SynLogic games, and forward the data through the base and tuned model to compute the distribution and corresponding KL. The reported forward KL is computed by taking macro average over all sequences in the calibration set.

\subsection{Sparsity}
\citet{mukherjee2025reinforcementlearningfinetunessmall, zhu2025pathnottaken} observed the different sparsity patterns in SFT and RLVR training, we try to understand if update patterns will have different sparsities under different offline learning objectives. Given a linear module $W_{\text{base}}$ from the base model and the corresponding module $W_{\text{trained}}$ in the trained model, $\epsilon$ be the sparsity threshold, the sparsity of the module is calculated as:
\begin{align*}
    \Delta W &= W_{\text{trained}} - W_{\text{base}},\\
\mathrm{sparsity}(W) 
&= \frac{1}{|\Delta W|}\sum_{i,j}\mathbf{1}\!\left(|\Delta W_{ij}| < \epsilon\right),
\end{align*}
A model's sparsity is calculated by taking a macro average across all its linear modules.

\subsection{Normalized Spectrum Shift}
Normalized Spectrum Shift (NSS) \citep{zhu2025pathnottaken} is a metric to measure the drift of the tuned model in the parameter space. For each module, we first perform singular value decomposition on the weight matrices to obtain the singular values, then measure the normalized distance between the singular value spaces.

\begin{align*}
    \text{NSS}(W)=\|\sigma(W_{+})-\sigma(W_0)\|_2/\|\sigma(W_0)\|_2
\end{align*}

\subsection{Gradient Rotation}
\label{grad-rotation}
We compute the rotation of gradients during the offline and online stage to understand if the offline and online training stages update the model in similar directions. Specifically, given the gradient if a module in the base model $\nabla_{\text{offline}} W$ in the offline stage and the gradient of the same module in the online stage  $\nabla_{\text{online}} W$, we first perform SVD on the two matrices to obtain $U_{\text{offline}}$ and $U_{\text{online}}$, then we evaluate the subspace rotation between the two matrices as follows:
\begin{align*}
    \cos_{\theta_i}(U):=\sigma_i(U_{\text{offline},k}\top U_{\text{online},k}), i=1,\dots, k
\end{align*}
where $U_K$ denotes and top-k subspace of $U$ and $k$ is equal to 128 in our case. To estimate the rotation of a module we simply take the average of the $\theta_i, \dots, \theta_k$. To estimate the rotation of model, we simply take macro average over all the linear modules.\\
In order to compute the gradients, we take a model that has been trained on offline data, and use another calibration set to run a pilot offline training on the model for 10 steps and collect the gradients. For the online gradient, we simply run GRPO on the model for 10 steps and collect the gradients. We take the means of the offline and online gradients across the steps and use the means to compute the module-wise rotations.

\begin{figure*}[!htbp]
    \centering
    \includegraphics[width=.5\linewidth]{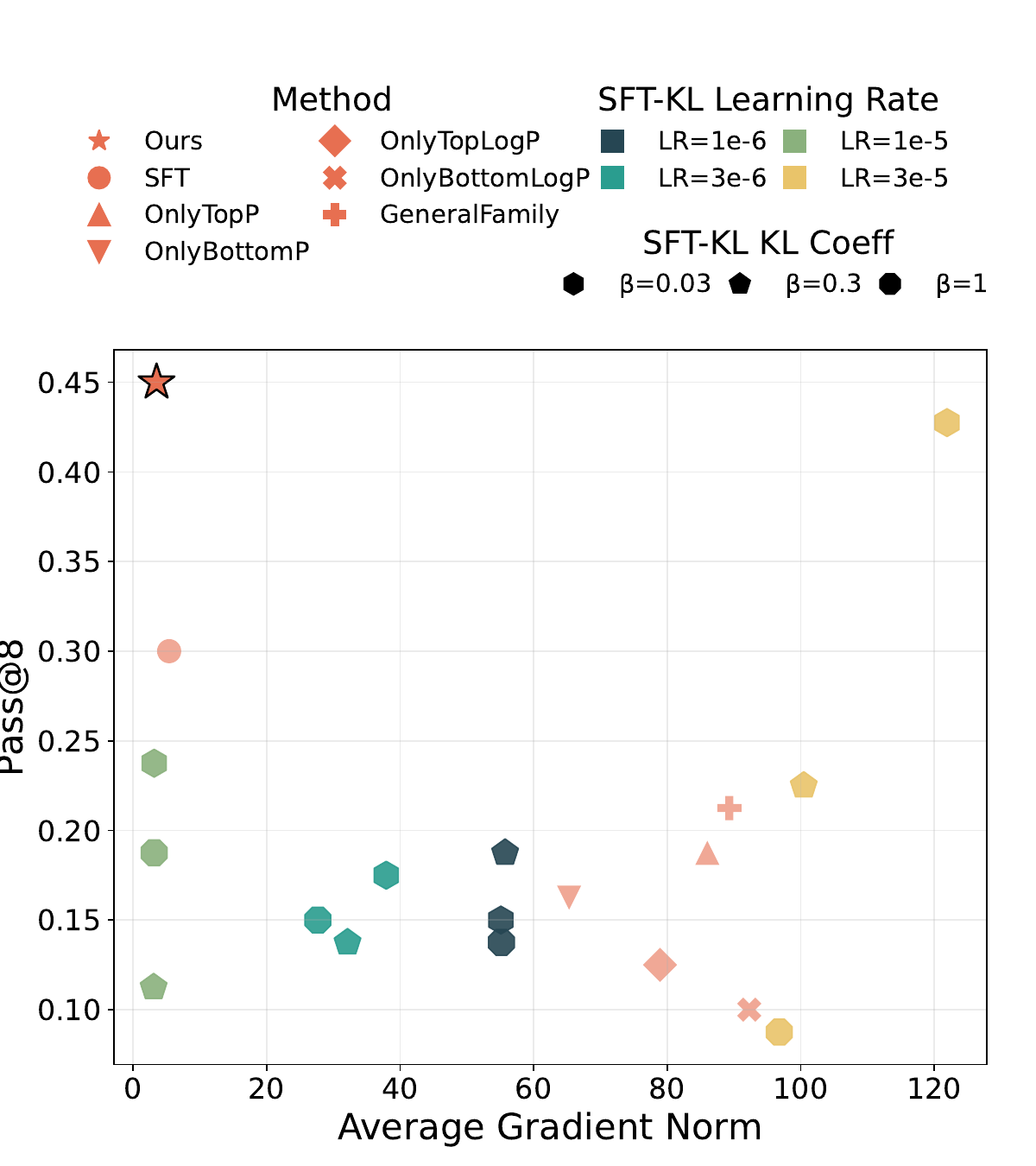}
    \caption{The comparison between different metrics versus SynLogic online pass@1. (a) offline model forward KL divergence against the base model. (b) offline model update sparsity against the base model. (c) average spectrum drift of different linear modules in the base and offline model. (d) average spectrum drift of different linear modules in the offline and online model.}
    \label{fig:metric-1-by-4}
\end{figure*}

\subsection{Results}
The results of metrics vs online pass@1 for Qwen3-1.7B-Base are visualized in Figure \ref{fig:metric-1-by-4}. From the figure, we can see that IS-SFT results in comparable forward KL divergence and much lower sparsity compared to SFT methods. Additionally, our method results in greater update during the offline stage, causing greater spectrum drift against the base model compared to SFT methods. In contrast, IS-SFT causes smaller updates during the online stage. 


\section{Suffix Change-of-Measure}
\label{app:suffix_com}
Let $x$ be a prompt and $y_{1:T}$ a token trajectory. For any $t\in\{1,\ldots,T\}$, define the prefix
$s_t \coloneqq (x,y_{<t})$ and the suffix $y_{t:T}$. For an autoregressive policy $\pi$,
\[
\pi(y_{t:T}\mid s_t)=\prod_{k=t}^T \pi(y_k\mid x,y_{<k}).
\]
Define the \emph{suffix likelihood ratio}
\[
\rho_{t:T}(x,y)\;\coloneqq\;\frac{\pi_\theta(y_{t:T}\mid s_t)}{\pi_\beta(y_{t:T}\mid s_t)}
=\prod_{k=t}^T \frac{\pi_\theta(y_k\mid x,y_{<k})}{\pi_\beta(y_k\mid x,y_{<k})}.
\]
Then for any measurable function $\varphi$ of the continuation (and the fixed prefix),
\begin{align*}
\mathbb{E}_{y_{t:T}\sim \pi_\theta(\cdot\mid s_t)}\!\big[\varphi(s_t,y_{t:T})\big]
&=\sum_{y_{t:T}} \pi_\theta(y_{t:T}\mid s_t)\,\varphi(s_t,y_{t:T})\\
&=\sum_{y_{t:T}} \pi_\beta(y_{t:T}\mid s_t)\,
\frac{\pi_\theta(y_{t:T}\mid s_t)}{\pi_\beta(y_{t:T}\mid s_t)}\,
\varphi(s_t,y_{t:T})\\
&=\mathbb{E}_{y_{t:T}\sim \pi_\beta(\cdot\mid s_t)}\!\big[\rho_{t:T}(x,y)\,\varphi(s_t,y_{t:T})\big],
\end{align*}
assuming $\pi_\beta(y_{t:T}\mid s_t)>0$ whenever $\pi_\theta(y_{t:T}\mid s_t)>0$.

\emph{Intuition.} Conditioning on the same prefix $s_t$, the two policies induce different distributions
over the remaining continuation $y_{t:T}$. The ratio $\rho_{t:T}$ is exactly the Radon--Nikodym
derivative that reweights $\pi_\beta$-suffix samples into unbiased expectations under $\pi_\theta$:
suffixes that are more likely under $\pi_\theta$ than $\pi_\beta$ receive larger weight, and vice versa.

\section{An Alternative Intuition: Suffix ratios as an off-policy estimate of return / value.}
\label{app:alt_intuition}
Recall the (discounted) terminal-feedback action-value under the \emph{target} policy:
\begin{align}
Q^{\pi_\theta}_\gamma(s_t,a_t)
\;\triangleq\;
\mathbb{E}_{a_{t+1:T}\sim \pi_\theta(\cdot\mid s_{t+1:T})}
\!\left[
\gamma^{\,T-t}\,R(\tau)
\;\middle|\;
s_t,a_t
\right],
\label{eq:q_def_recall}
\end{align}
where $\tau=(s_1,a_1,\ldots,s_T,a_T)$ and $R(\tau)$ is observed only at $T$.

\emph{Change of measure on the continuation.}
Condition on the same prefix decision $(s_t,a_t)$ and rewrite the $\pi_\theta$-continuation expectation
as an expectation over continuations sampled from the logging policy $\pi_\beta$:
\begin{align}
Q^{\pi_\theta}_\gamma(s_t,a_t)
&=
\sum_{a_{t+1:T}}
\pi_\theta(a_{t+1:T}\mid s_t,a_t)\;\gamma^{T-t}R(\tau)
\nonumber\\
&=
\sum_{a_{t+1:T}}
\pi_\beta(a_{t+1:T}\mid s_t,a_t)\;
\underbrace{\frac{\pi_\theta(a_{t+1:T}\mid s_t,a_t)}{\pi_\beta(a_{t+1:T}\mid s_t,a_t)}}_{w_{t+1:T}(\tau)}
\;\gamma^{T-t}R(\tau)
\nonumber\\
&=
\mathbb{E}_{\tau\sim \pi_\beta}
\!\left[
\gamma^{\,T-t}\,R(\tau)\,w_{t+1:T}(\tau)
\;\middle|\;
s_t,a_t
\right],
\label{eq:q_pdis_suffix}
\end{align}
with the suffix importance ratio
\[
w_{t+1:T}(\tau)\;\triangleq\;\prod_{j=t+1}^T \frac{\pi_\theta(a_j\mid s_j)}{\pi_\beta(a_j\mid s_j)}.
\]
\emph{Intuition:} $w_{t+1:T}$ “translates” logged suffixes into what $\pi_\theta$ would typically see.
If the logged continuation is unlikely under $\pi_\theta$, it should contribute little to $\pi_\theta$’s
expected return-to-go from $(s_t,a_t)$.

\paragraph{From $Q$ to a token-level return estimator.}
Given one logged trajectory $\tau\sim\mathcal{D}$, a single-sample plug-in estimator of
\eqref{eq:q_pdis_suffix} is exactly your per-token credit weight
\begin{align}
G_t(\tau)\;\triangleq\;\gamma^{\,T-t}R(\tau)\,w_{t+1:T}(\tau),
\qquad
\text{so that}\quad
G_t(\tau)\approx Q^{\pi_\theta}_\gamma(s_t,a_t).
\label{eq:G_as_Qhat}
\end{align}

\emph{Intuition:} uniform SFT corresponds to replacing $Q$ by a constant (every token gets equal credit),
while \name replaces it by an outcome-aware return estimate that (i) propagates terminal feedback back
to earlier decisions (via $\gamma^{T-t}R$) and (ii) discounts suffixes that $\pi_\theta$ would not
actually realize during on-policy rollouts (via $w_{t+1:T}$).

\section{Details On Baselines}
\label{app:baselines}

\subsection{Token-Adaptive Loss Reweighting (TALR)}
\label{app:talr_hparams}

TALR reweights token-level negative log-likelihood (NLL) by an exponential function of token difficulty.
Given token probability $p_t$ for the supervised token at position $t$, define token NLL
\[
\ell_t = -\log p_t.
\]
TALR assigns an adaptive weight
\[
\tilde{w}_t = \exp\!\left(-\frac{\ell_t}{\tau}\right),
\qquad
w_t = \max\!\big(\mathrm{sg}(\tilde{w}_t),\, w_{\min}\big),
\]
where $\mathrm{sg}(\cdot)$ denotes \emph{stop-gradient} (weights treated as constants in backprop).

The reweighted batch loss is the (token-)mean:
\[
\mathcal{L}_{\textsc{talr}} = \frac{1}{N}\sum_{t=1}^{N} w_t\,(-\log p_t),
\]
with $N$ the number of supervised tokens in the batch.

\paragraph{Key TALR hyperparameters.}
\begin{itemize}
    \item \textbf{Weight floor} $w_{\min}$: fixed to $0.01$ in all experiments (prevents vanishing weights on very hard tokens).
    \item \textbf{Temperature} $\tau$: selected \textbf{dynamically} each step as the \textbf{median of the average sequence loss within the batch}.
\end{itemize}

\subsection{Beyond Log Likelihood}
\label{app:bll-hparams}

Here we summarize the hyper-parameter settings reported in
\citet{li2025beyond}.

The base learning rate is
$5\times 10^{-5}$.

\paragraph{Objective hyperparameters.}
Their core parametric family is
$f_{\alpha}(p)=\frac{1-p^{\alpha}}{\alpha}$ (with $\alpha\to 0$ recovering NLL).
In the main math results, they instantiate several concrete choices, including:
(i) NLL $-\log p$,
(ii) $-p$ (equivalently the $\alpha=1$ member up to an additive constant),
and (iii) a hard-thresholded NLL of the form $-\log(p)\cdot\mathbb{I}[p\ge 0.2]$.
They also discuss higher-power prior-leaning variants (e.g., $\alpha=10$).

\begin{table}[t]
\small
\centering
\renewcommand{\arraystretch}{1.15}
\begin{tabular}{l l l l}
\toprule
\textbf{Key} & \textbf{Name} & \textbf{Per-token loss $f(p)$} & \textbf{Hyperparameters / mask} \\
\midrule
\texttt{original} &
NLL (standard SFT) &
$-\log p$ &
None \\
\texttt{GeneralFamily-$\alpha$} &
Probability family &
$\displaystyle \frac{1-p^{\alpha}}{\alpha}$ &
$\alpha$ (with $\alpha\to 0$ recovering $-\log p$) \\
\texttt{p} &
Plain-$p$ objective &
$1-p$ &
None (equiv. to \texttt{GeneralFamily-1} up to constants) \\
\texttt{OnlyTopP-$q$} &
Top-thresholded (plain-$p$) &
$(1-p)\,\mathbf{1}[p\ge q]$ &
$q\in[0,1]$ \\
\texttt{OnlyBottomP-$q$} &
Bottom-thresholded (plain-$p$) &
$(1-p)\,\mathbf{1}[p\le q]$ &
$q\in[0,1]$ \\
\texttt{OnlyTopLogP-$q$} &
Top-thresholded (NLL) &
$-\log(p)\,\mathbf{1}[p\ge q]$ &
$q\in[0,1]$ \\
\texttt{OnlyBottomLogP-$q$} &
Bottom-thresholded (NLL) &
$-\log(p)\,\mathbf{1}[p\le q]$ &
$q\in[0,1]$ \\
\midrule
\multicolumn{4}{l}{\textit{Paper-only (used for analysis/ablations; not exposed as repo keys)}}\\
\bottomrule
\end{tabular}
\caption{
Objectives in \emph{Beyond Log Likelihood}. Here
$p := p_\theta(y_t\mid y_{<t},x)$ denotes the model probability of the ground-truth token at step $t$,
and the sequence loss is $\sum_t f(p_t)$.
}
\label{tab:bll_objectives}
\end{table}
\section{Discussion: Should Offline Training Match Online RL Characteristics In The Two-Stage Process?}
\label{app:discussion_twostage}
\begin{figure}
\centering
  \begin{subfigure}[t]{0.49\linewidth}
    \centering
    \includegraphics[width=\linewidth]{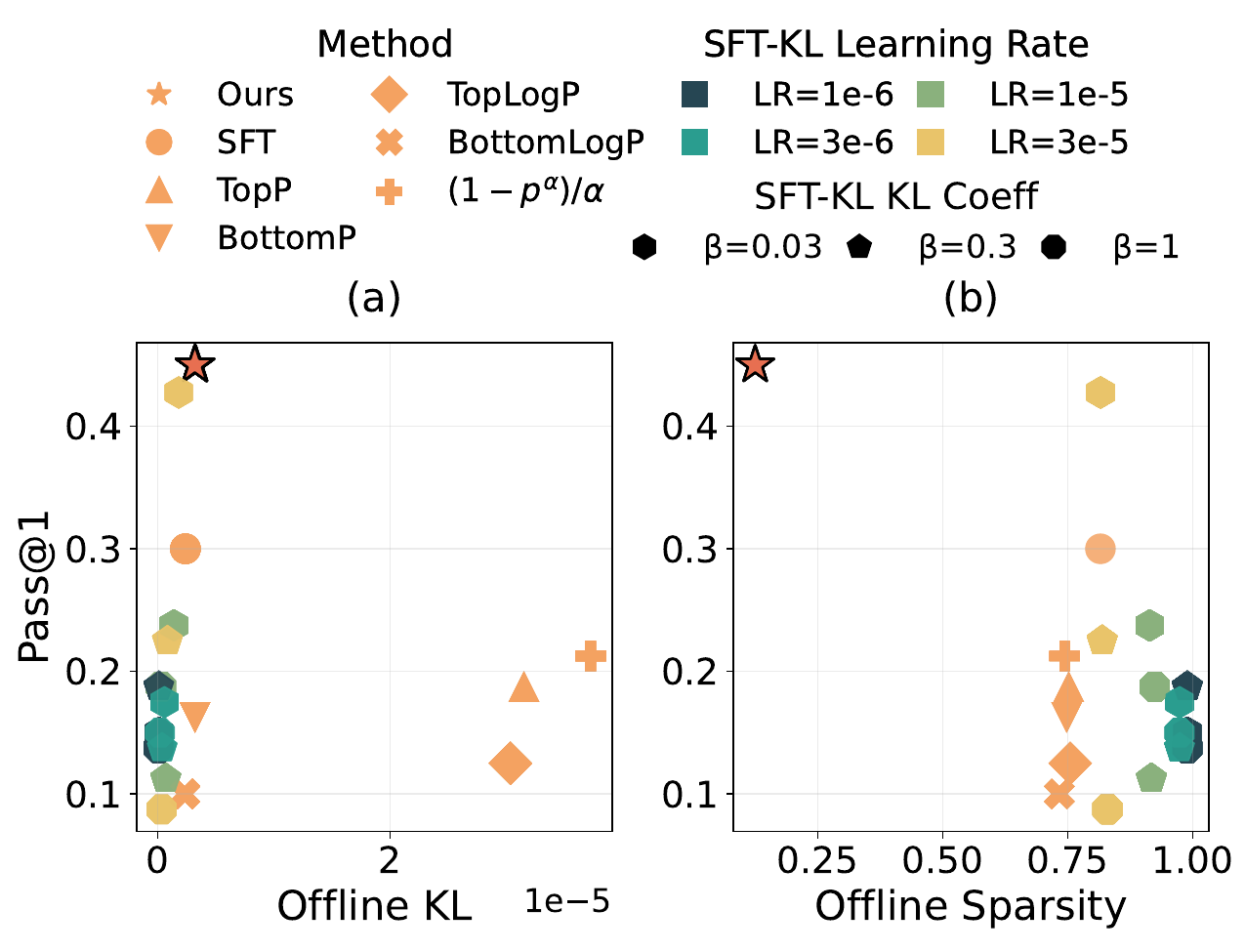}
    \caption{KL-to-base and update sparsity of offline updates versus Pass@1.}
    \label{fig:kl_appx_pass1}
  \end{subfigure}
  \begin{subfigure}[t]{0.49\linewidth}
    \centering
    \includegraphics[width=\linewidth]{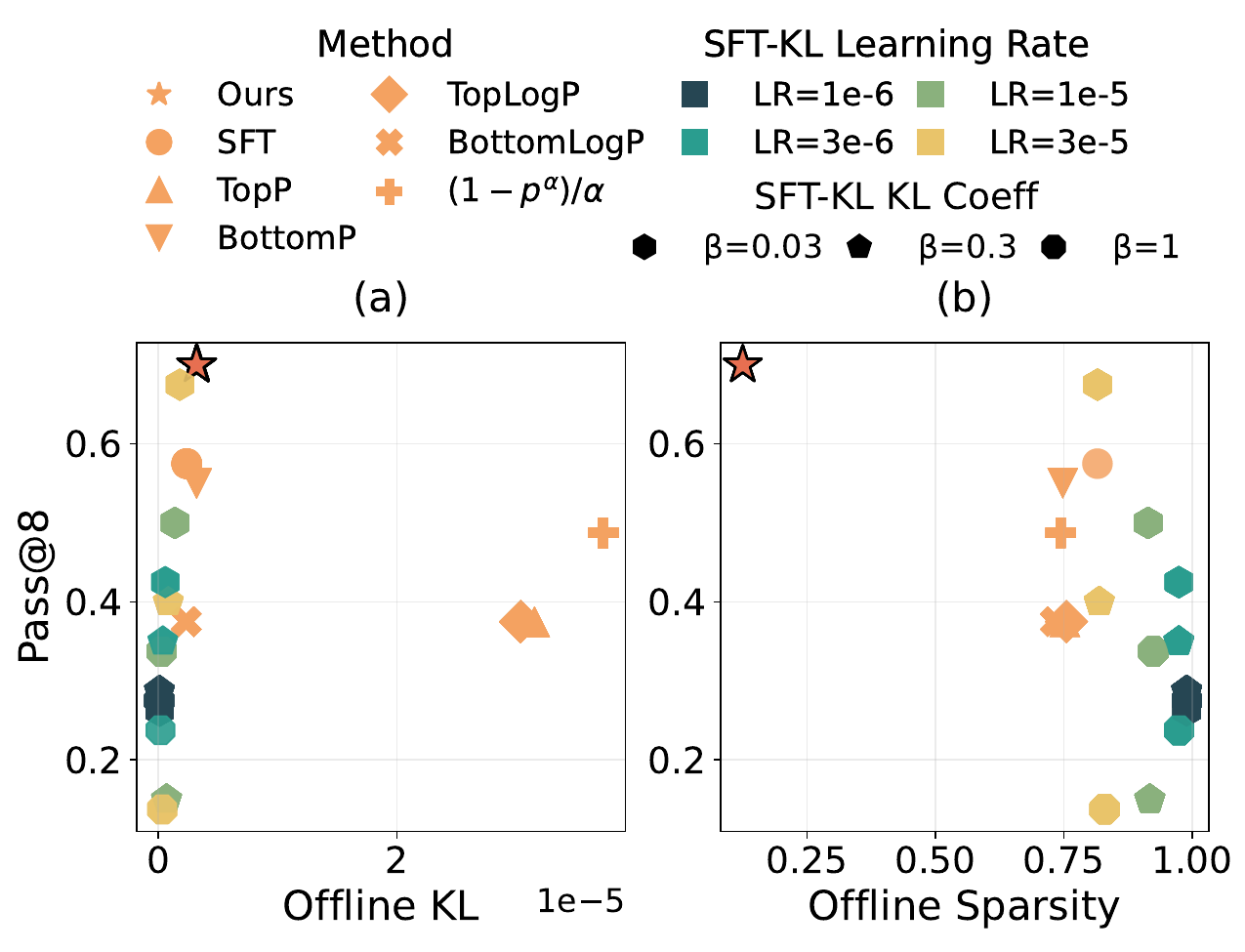}
    \caption{KL-to-base and update sparsity of offline updates versus Pass@8.}
    \label{fig:kl_appx_pass8}
  \end{subfigure}
    \caption{Offline update strength measured by KL to base model and sparsity of parameter updates.}
    \label{fig:kl_appx}
\end{figure}

While it is tempting to treat common stabilization signals—smaller KL to the base~\citep{shenfeld2025rlrazor} policy, sparser~\citep{mukherjee2025reinforcementlearningfinetunessmall} (lower-magnitude) updates, or a smaller ``rotation''~\citep{zhu2025pathnottaken} away from the base representation, these quantities primarily measure \emph{conservatism}, not actually an accurate ``mismatch correction''. 

Figure~\ref{fig:model_characteristics}-a shows that KL penalties ensure smaller drift yet does not boost the performance compared with vanilla SFT, and while \name leads to more aggressive KL drifts and denser updates~\ref{fig:model_characteristics}-b, it performs better. In fact, for downstream RL the goal is not to minimize movement per se, but to move in the \emph{right directions}: toward behaviors that improve expected return under on-policy rollouts, even if that requires nontrivial deviation from the base. Consequently, a checkpoint that “looks stable” by these proxies may still yield weak post-RL gains (or learn slowly) because it has not acquired the right inductive biases, coverage, or credit-assignment structure that makes subsequent RL compute-efficient.

\section{Table For Offline vs Online Metrics (Pass@1 and Pass@8)}
\label{app:offline_online_tables}
\begin{figure*}[t]
    \centering
    \includegraphics[width=\linewidth]{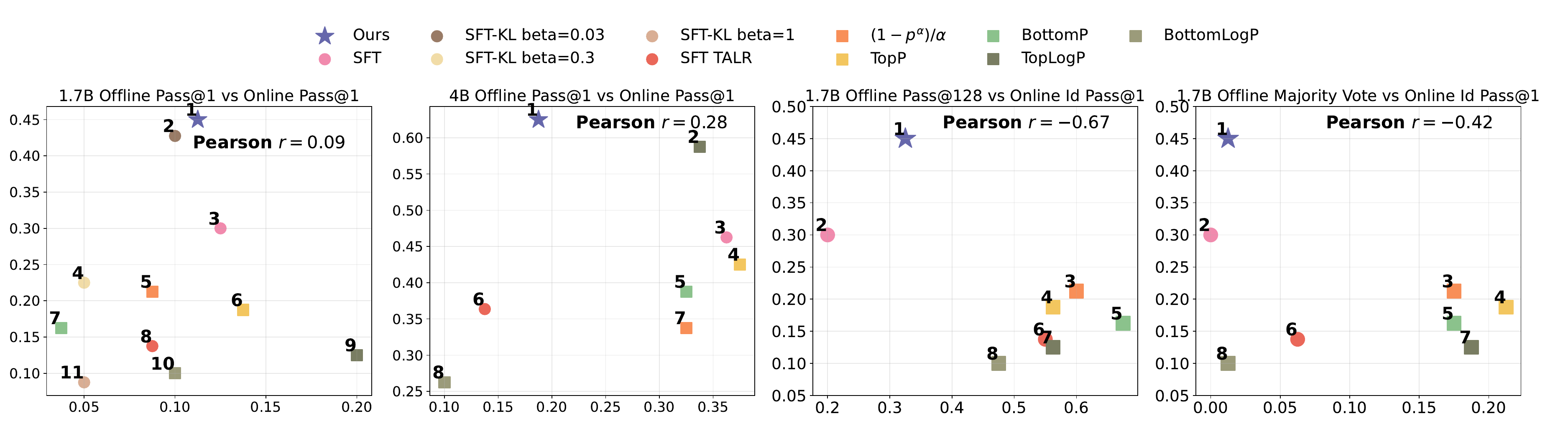}
    \caption{Visualization of offline vs online performance. \textbf{(a)}: Qwen3-1.7B-Base-Base offline pass@1 versus online pass@1. \textbf{(b)}: Qwen3-4B-Base offline pass@1 versus online pass@1. \textbf{(c)}: Qwen3-1.7B-Base offline pass@128 versus online pass@1. \textbf{(d)}: Qwen3-1.7B-Base offline majority vote versus online pass@1.}
    \label{fig:off-on-curve}
\end{figure*}

One surprising finding we made is that stronger offline performance does not necessarily lead to stronger post-RL performance. The results are visualized in Figure \ref{fig:off-on-curve}. This section shows the detailed offline and online performances of Qwen3-1.7B-Base and Qwen3-4B-Base trained with different learning objectives. From the figures we can clearly see the line segments intersect with each other, and that offline-online performances do not have a consistent ranking. Detailed evaluation results of Qwen3-1.7B-Base and Qwen3-4B-Base can be found in Table~\ref{tab:1.7b-eval} and \ref{tab:4b-eval}, respectively. 

\begin{table}[h!]
    \centering
    \begin{tabular}{lccccccc}
        \toprule
        \multirow{2}{*}{Objective} 
         & \multicolumn{4}{c}{Offline} & \multicolumn{2}{c}{Online} \\
         \cmidrule{2-5} \cmidrule{6-7}
         & Pass@1 & Pass@8 & Pass@128 & maj vote & Pass@1 & Pass@8 \\
         \midrule
         \name             & 32.50\%        & 37.50\%        & 31.25\%          & 1.25\%           & 22.50\%       & 56.25\%       \\
\name             & 20.00\%        & 23.75\%        & 32.50\%          & 1.25\%           & 45.00\%       & 70.00\%       \\
SFT            & 22.50\%        & 27.50\%        & 20.00\%          & 0.00\%           & 30.00\%       & 57.50\%       \\
SFT TALR       & 8.75\%         & 35.00\%        & 55.00\%          & 6.25\%           & 13.75\%       & 48.75\%       \\
SFT-KL($\beta=0.03$)         & 10.00\%        & 35.00\%        &        -         &           -       & 42.75\%       & 67.50\%       \\
SFT-KL($\beta=0.3$)           & 5.00\%         & 40.00\%        &        -          &           -       & 22.50\%       & 40.00\%       \\
SFT-KL($\beta=1.0$)          & 5.00\%         & 28.75\%        &         -         &           -       & 8.75\%        & 13.75\%       \\
TopP       & 13.75\%        & 33.75\%        & 56.25\%          & 21.25\%          & 18.75\%       & 37.50\%       \\
TopLogP    & 20.00\%        & 37.50\%        & 56.25\%          & 18.75\%          & 10.00\%       & 37.50\%       \\
BottomP    & 3.75\%         & 30.00\%        & 56.25\%          & 13.75\%          & 16.25\%       & 55.00\%       \\
BottomLogP & 10.00\%        & 28.75\%        & 47.50\%          & 1.25\%           & 10.00\%       & 37.50\%       \\
$(1-p)^\alpha/\alpha$  & 8.75\%         & 27.50\%        & 60.00\%          & 17.50\%          & 21.25\%       & 48.75\%  \\
         \bottomrule
    \end{tabular}
    \caption{Offline and online pass rates of Qwen3-1.7B-Base with different learning objectives. All results are evaluated on SynLogic.}
    \label{tab:1.7b-eval}
\end{table}

\begin{table}[h!]
    \centering
    \begin{tabular}{lccccccc}
        \toprule
        \multirow{2}{*}{Objective} 
         & \multicolumn{2}{c}{Offline} & \multicolumn{2}{c}{Online} \\
         \cmidrule{2-3} \cmidrule{4-5}
         & Pass@1 & Pass@8 & Pass@1 & Pass@8 \\
         \midrule
\name           & 32.50\% & 37.50\% & 63.75\% & 76.25\% \\
SFT            & 10.00\% & 26.25\% & 42.50\% & 71.25\% \\
SFT TALR       & 13.75\% & 45.00\% & 36.40\% & 68.80\% \\
TopP       & 37.50\% & 45.00\% & 42.50\% & 63.75\% \\
TopLogP    & 33.75\% & 52.50\% & 58.76\% & 72.50\% \\
BottomP    & 32.50\% & 60.00\% & 38.75\% & 71.25\% \\
BottomLogP & 10.00\% & 16.25\% & 26.25\% & 42.50\% \\
$(1-p)^\alpha/\alpha$  & 32.50\% & 47.50\% & 33.75\% & 63.75\% \\
    \bottomrule
    \end{tabular}
    \caption{Offline and online pass rates of Qwen3-4B-Base with different learning objectives. All results are evaluated on SynLogic.}
    \label{tab:4b-eval}
\end{table}


\end{document}